\documentclass[acmsmall,screen,review=false]{acmart}

\AtBeginDocument{%
  \providecommand\BibTeX{{%
    \normalfont B\kern-0.5em{\scshape i\kern-0.25em b}\kern-0.8em\TeX}}}

\usepackage{balance,booktabs,stmaryrd,listings,multirow,wrapfig,graphicx,caption,dashbox,pifont,xcolor,colortbl,color,soul,enumitem}
\usepackage{lscape}
\usepackage{wasysym}
\usepackage{tikz}
\usepackage{booktabs}
\usepackage{bm}
\usepackage[ruled,linesnumbered,vlined]{algorithm2e}

\newtheorem{principle}{Rule}

\setcopyright{acmcopyright}
\copyrightyear{2018}
\acmYear{2018}
\acmDOI{10.1145/1122445.1122456}

\acmJournal{JACM}
\acmVolume{37}
\acmNumber{4}
\acmArticle{111}
\acmMonth{8}

\begin{document}

\title{Indexing Context-Sensitive Reachability}

\author{Qingkai Shi}
\affiliation{%
  \institution{Ant Group}
  \country{China}}
\email{qingkai.sqk@antgroup.com}

\author{Yongchao Wang}
\affiliation{%
  \institution{The Hong Kong University of Science and Technology and Ant Group}
  \country{China}}
\email{ywanghz@cse.ust.hk}

\author{Charles Zhang}
\affiliation{%
  \institution{The Hong Kong University of Science and Technology}
  \city{Hong Kong}
  \country{China}}
\email{charlesz@cse.ust.hk}

\begin{abstract}
  
Many context-sensitive data flow analyses can be formulated as 
a variant of the all-pairs Dyck-CFL reachability problem,
which, in general, is of sub-cubic time complexity and quadratic space complexity.
Such high complexity significantly limits the scalability of context-sensitive data flow analysis
and is not affordable for analyzing large-scale software.
This paper presents \textsc{Flare}, a reduction
from the CFL reachability problem to the conventional graph reachability problem for context-sensitive data flow analysis.
This reduction
allows us to benefit from recent advances in reachability indexing schemes,
which often consume almost linear space for answering reachability queries in almost constant time.
We have applied our reduction to a context-sensitive alias analysis and a context-sensitive information-flow analysis for C/C++ programs.
Experimental results 
on standard benchmarks and open-source software
demonstrate that we can achieve orders of
magnitude speedup at the cost of only moderate space to store the indexes.
The implementation of our approach is publicly available.
\end{abstract}



\maketitle

\section{Introduction}
\label{sec:intro}

The context-free language (CFL) reachability problem is a generalization of the conventional graph reachability problem \cite{yannakakis1990graph}. A vertex $v$ is CFL-reachable from a vertex $u$ if and only if there is a path from the vertex $u$ to the vertex $v$, and the string of the edge labels on the path follows a given context-free grammar.
CFL reachability has been broadly used in program analysis for a wide range of applications, including
context-sensitive data flow analysis~\cite{reps1995precise}, 
program slicing~\cite{reps1994speeding}, 
shape analysis~\cite{reps1995shape},
type-based flow analysis~\cite{rehof2001type, kodumal2004set, pratikakis2006existential,milanova2020flowcfl},
pointer analysis \cite{sridharan2005demand, sridharan2006refinement, pratikakis2006existential, zheng2008demand, xu2009scaling, yan2011demand, shang2012demand, zhang2013fast, zhang2014efficient}, and
debugging~\cite{cai2018calling},
to name just a few.

\begin{figure*}[t]
	\centering
	\includegraphics[width=\columnwidth]{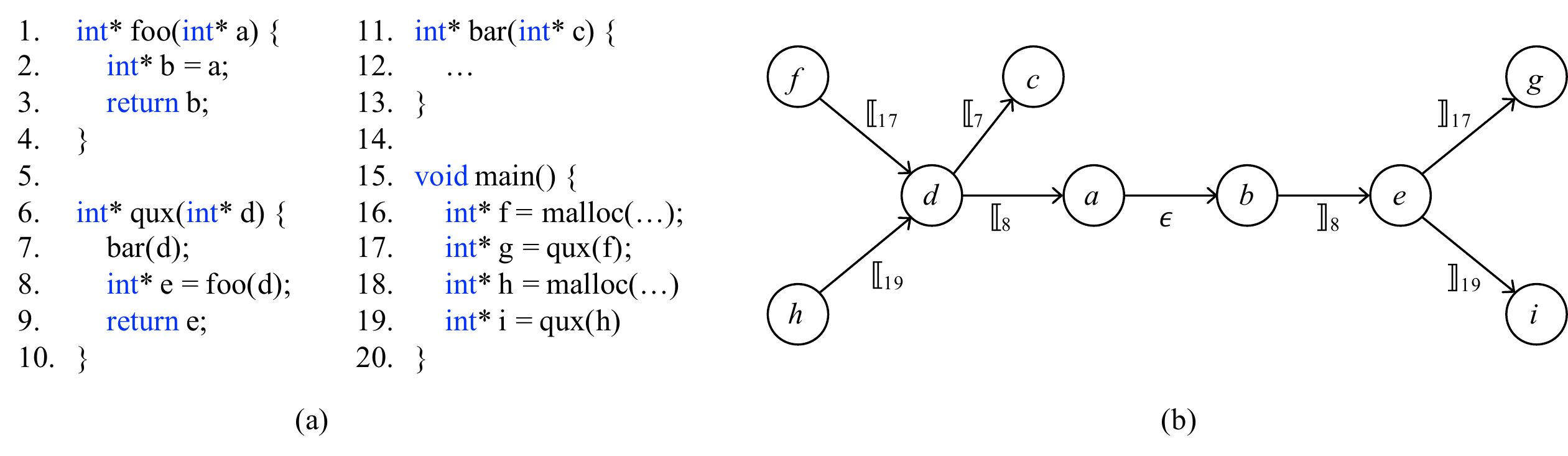}
	\caption{(a) Code for illustration. (b) Inter-procedural data-dependence graph, where each directed edge represents a data-dependence relation in the code. The parentheses $\llbracket_i$ and $\rrbracket_i$ stand for the call and return operations at Line $i$.}
	\label{fig:intro_ex}
\end{figure*}

This paper focuses on the problem of context-sensitive data flow analysis,
where an extended Dyck-CFL is used to capture the paired call and return using matched parentheses.
We use an ``extended'' Dyck-CFL because the standard one fails to capture many valid data flows containing partially matched parentheses~\cite{kodumal2004set}.
Intuitively, the extended Dyck-CFL includes all sub-strings of a standard Dyck word.
For instance, 
in the inter-procedural data-dependence graph in Figure \ref{fig:intro_ex}, 
the vertex $i$ is context-sensitively reachable from the vertices $b$ because the string of the edge labels, $\rrbracket_8\rrbracket_{19}$, does not contain any mismatched parentheses and, thus,
is a sub-string of a standard Dyck word like $\llbracket_{19}\llbracket_8\rrbracket_8\rrbracket_{19}$.
In contrast, 
the vertex $i$ is not context-sensitively reachable from the vertex $f$ because the string of the edge labels, $\llbracket_{17}\llbracket_8\rrbracket_8\rrbracket_{19}$, contains mismatched parentheses.
To distinguish from the standard Dyck-CFL reachability problem,
we refer to the extended version as the context-sensitive reachability (CS-reachability) problem as it is specially used in the context-sensitive data flow analysis.\footnote{By context-sensitive reachability, we do not mean the underlying language is a context-sensitive language but still a context-free language specially used for context-sensitive data flow analysis.}

Recently, some fast algorithms have been proposed to address the standard Dyck-CFL reachability problems on a few special graphs, such as trees~\cite{zhang2013fast, yuan2009efficient}, bidirected graphs~\cite{zhang2013fast, chatterjee2017optimal},
and graphs of constant tree-width~\cite{chatterjee2017optimal}.
Nevertheless, 
due to the differences in the underlying CFLs and graph structures, these approaches cannot be directly employed in context-sensitive data flow analysis.
In practice, for context-sensitive alias analysis~\cite{li2013precise,li2011boosting}, information-flow analysis~\cite{lerch2014flowtwist,arzt2014flowdroid},
and all other IFDS-based data flow analyses,
answering a CS-reachability query still relies on the typical tabulation algorithm~\cite{reps1995precise,reps1994speeding},
either in an exhaustive manner or in a demand-driven fashion.
The exhaustive manner computes a transitive closure,
which is of at least quadratic complexity and is unaffordable for large-scale software.
The demand-driven manner traverses the graph for every reachability query and, thus, is not efficient at responding to a query.

This paper proposes indexing schemes for solving the all-pairs CS-reachability problem,
so that we can efficiently tell the CS-reachability relation between any pair of vertices without computing an expensive transitive closure or performing a full graph traversal.
Our key insight is that the CS-reachability problem can be reduced to a conventional reachability problem within linear time and space, by building a special graph structure we refer to as the indexing graph.
Thanks to the recent advances in the field of graph database~\cite{jin2011path,yildirim2010grail,cheng2013tf,wang2006dual,jin20093}, the reduction allows us to employ existing indexing schemes for conventional graph reachability
to significantly speed up CS-reachability queries, at the cost of only a moderate space overhead.

We have implemented a tool, namely \textsc{Flare}, to build the indexing graph for a given context-sensitive analysis, so that the CS-reachability problem can be reduced to the conventional graph reachability problem.
Based on the reduction, we then apply two different existing indexing schemes for speeding up context-sensitive information-flow analysis and context-sensitive alias analysis, respectively.
In the evaluation,
we conducted experiments on twelve standard benchmark programs and four open-source systems to measure the time cost for building indexes, the space cost for storing the indexes, and the query time using the indexes.
We also compared our method to a few baseline approaches,
which showed that we can achieve orders of magnitude speedup for answering an alias or information-flow query with only a moderate overhead to build and store the indexes.
In summary, the principal contributions of this paper are three-fold and listed as follows:
\begin{itemize}
	\item We propose a reduction of linear time and space complexity from the CS-reachability problem to the conventional graph reachability problem.
	We prove its correctness and analyze its time and space complexity.
	\smallskip
	
	\item We present two typical applications of our reduction, namely context-sensitive information-flow analysis and context-sensitive alias analysis.
	Through the two applications,
	we also summarize the criteria of selecting a proper indexing scheme in practice.
	\smallskip
	
	\item We evaluate the time and the space overhead of building the indexes,
	and compare our method to existing techniques. 
	The results showed orders of magnitude speedup for answering CS-reachability queries with just a moderate space overhead.
\end{itemize}


\section{Background}
\label{sec:preliminaries}

In this section,
we review the background of context-sensitive reachability (Section~\ref{subsec:csr}) as well as existing indexing schemes for conventional graph reachability (Section~\ref{subsec:gr}).
We also discuss the connections and the gaps between context-sensitive reachability and existing reachability indexing schemes (Section~\ref{subsec:gaps}).

\subsection{Context-Sensitive Reachability}
\label{subsec:csr}

In this paper, we study the all-pairs context-sensitive reachability problem on various flow graphs of a program.
These graphs include the program dependence graph~\cite{ferrante1987program}, the value-flow graph~\cite{cherem2007practical,sui2014detecting},
the exploded super graph~\cite{reps1995precise}, and many others.
Generally, these graphs can be uniformly defined as a program-valid graph, which captures the modular program structure~\cite{chatterjee2017optimal}.

\begin{definition}[Program-Valid Graph]
	Given	
	an alphabet $\Sigma_k=\{\epsilon\}\cup\{\llbracket_i, \rrbracket_i\}_{i=1}^k$,
	a program-valid graph $G$ is a $\Sigma_k$-labeled directed graph that can be partitioned to sub-graphs such that every sub-graph has only $\epsilon$-labeled edges, and there exists a constant $\alpha\ge0$ such that every sub-graph has $\alpha$ or fewer vertices with $\llbracket_i$-labeled incoming edges or $\rrbracket_i$-labeled outgoing edges.
\end{definition}

Intuitively speaking, every sub-graph in a program-valid graph represents the local graph of a function. The constant $\alpha$ indicates that every function in a program has only a few function parameters and return values.
For example, the inter-procedural data-dependence graph in Figure~\ref{fig:intro_ex} is program-valid because it can be partitioned into four parts, $\{a, b\}$, $\{c\}$, $\{d, e\}$, and $\{f, g\}$.
The four parts correspond to the four functions, \textsf{foo}, \textsf{bar}, \textsf{qux}, and \textsf{main}, respectively. Each part has at most two vertices with parenthesis-labeled incoming or outgoing edges, which stand for the function call and return operations in the program.
From now on, given a program-valid graph,
we use $V$ to represent the vertex set and $E\subseteq V\times V$ the edge set. 
Given any edge $(u, v)\in E$, $L(u, v) \in \Sigma_k$ returns the edge label.

\begin{definition}[Context-Sensitive Reachability]
	Given two vertices $v_0$ and $v_m$ on a program-valid graph,
	we say the vertex $v_m$ is context-sensitively reachable (or CS-reachable) from the vertex $v_0$
	if and only if there is a path $(v_0, v_1, v_2, \ldots, v_m)$ on the graph
	such that the concatenation of the edge labels,
	$L(v_0, v_1)L(v_1, v_2)\ldots L(v_{m-1}, v_m)$, can be derived from the start symbol $S$ of the context-free grammar in Figure~\ref{fig:intro_grammar}.
\end{definition}

\begin{figure*}[t]
	\small
	\centering
	\begin{tabular}{rcl}
		$S$&$\enskip\rightarrow\enskip$& $P\enskip N$
		\\
		$P$&$\enskip\rightarrow\enskip$& $M\enskip P\enskip|\enskip\rrbracket_i\enskip P\enskip|\enskip\epsilon$
		\\
		$N$&$\enskip\rightarrow\enskip$& $M\enskip N\enskip|\enskip\llbracket_i\enskip N\enskip|\enskip\epsilon$
		\\
		$M$&$\enskip\rightarrow\enskip$& $\llbracket_i\enskip M\enskip\rrbracket_i\enskip|\enskip M\enskip M\enskip|\enskip\epsilon$
	\end{tabular}
	\caption{The context-free grammar of an extended Dyck-CFL, which is defined on the alphabet $\Sigma_k=\{\epsilon\}\cup\{\llbracket_i, \rrbracket_i\}_{i=1}^k$ for achieving context-sensitivity \cite{kodumal2004set}. The grammar for the standard Dyck-CFL only has the last production, $M$, that produces matched parentheses.}
	\label{fig:intro_grammar}
\end{figure*}

By definition,
the context-free grammar in Figure~\ref{fig:intro_grammar}
allows three kinds of CS-reachable paths on the program-valid graph:
\begin{enumerate}
	\item $P$-paths: paths whose edge-label strings can be derived from the symbol $P$ of the grammar.
	By definition,
	a parenthesis on a $P$-path is either a right-parenthesis or correctly matched.
	In a program analysis, a $P$-path often represents the propagation of a data-flow fact from a callee function to a caller function. For instance,
	the path $(b,e,g)$ in Figure~\ref{fig:intro_ex} is a $P$-path.
	\smallskip
	
	\item $N$-paths: paths whose edge-label strings can be derived from the symbol $N$ of the grammar. 
	By definition,
	a parenthesis on an $N$-path is either a left-parenthesis or correctly matched.
	In a program analysis, an $N$-path often represents the propagation of a data-flow fact from a caller function to a callee function. For instance,
	the path $(f, d, a, b)$ in Figure~\ref{fig:intro_ex} is an $N$-path.
	\smallskip
	
	\item $PN$-paths: the concatenation of $P$-paths and $N$-paths, which implies that a data-flow fact returned from a callee function
	is passed again to a callee function.
\end{enumerate}

Hence, to answer a CS-reachability query, we in fact need to check if there is a $P$-path, $N$-path, or $PN$-path between two vertices. To this end, the state-of-the-art method is to employ \citet{reps1994speeding,reps1995precise}'s tabulation algorithm, which has been used in a wide range of applications, including alias analysis~\cite{li2013precise,li2011boosting}, information-flow analysis~\cite{lerch2014flowtwist,arzt2014flowdroid}, as well as all other data flow analyses built on top of the IFDS framework~\cite{reps1995precise}.

Algorithm~\ref{alg:tabulation} illustrates the spirit of the tabulation algorithm.
Basically, to answer a CS-reachability query,
the algorithm performs a depth-first graph traversal over the input program-valid graph but, during the traversal, employs the summary edges to avoid repetitively visiting a function.
As defined in Definition~\ref{def:se},
a summary edge tabulates an input and an output of a function.
\citet{reps1994speeding} showed that the number of summary edges is bounded by $O(\alpha^2|V|)$ and the time to build all summary edges is bounded by $O(\alpha|E| + \alpha^3|V|)$.

As an example of the summary edges,
when traversing the inter-procedural data-dependence graph in Figure~\ref{fig:intro_ex},
we can add a summary edge from the vertex $d$ to the vertex $e$.
This summary edge allows us to skip the function \textsf{qux} whenever a graph traversal reaches the vertex $d$.
Since we never visit a function more than once, answering a CS-reachability query using the tabulation algorithm is of linear complexity with respect to the graph size and the number of summary edges.

\begin{definition}[Summary Edge and Summary Path]\label{def:se}
	A summary edge $(v_0, v_m)$ is an extra edge added to the program-valid graph $G=(V, E)$ such that
	$v_0, v_m \in V$ and there is a path $(v_0, v_1, v_2, \ldots, v_m)$, which we refer to as a summary path,
	such that 
	$L(v_0, v_1)L(v_{m-1}, v_m)=\llbracket_i\rrbracket_i$,
	and the label string $=L(v_1, v_2)L(v_2, v_3)\ldots L(v_{m-2}, v_{m-1})$ can be derived from the symbol $M$ of the grammar in Figure~\ref{fig:intro_grammar}.
\end{definition}

While the tabulation algorithm avoids repetitively visiting a function when answering a CS-reachability query,
it is not efficient for frequent CS-reachability queries because we need to traverse the graph for every query.
To expedite CS-reachability queries,
the usual manner is to build a transitive closure so that we can answer each query in constant time.
However,
building a transitive closure is notoriously expensive (at least quadratic complexity), which is unaffordable for large-scale graphs.
To resolve the dilemma between traversing the graph for every query and computing an expensive transitive closure,
this paper proposes a novel use of the summary edges, which allows us to answer each CS-reachability query within ``almost'' constant time via the indexing schemes for conventional graph reachability.

\begin{algorithm}[t]\small
  \caption{Query if the target vertex $t$ is CS-reachable from the source vertex $s$.}
  \label{alg:tabulation}
  \SetKwFunction{Q}{Query}
  \SetKwFunction{T}{Tabulate}
  \SetKwProg{Proc}{Procedure}{}{}
  $E^s = $ \{~a set of summary edges pre-computed using \citet{reps1994speeding}'s method~\}\;
  \Proc{\Q{$G=(V, E)$, $s\in V$, $t\in V$}}{
    \textbf{if} {$s$ has been visited before by this procedure} \textbf{then} \textbf{return} {false}\;
    \textbf{if} {$s=t$} \textbf{then} \textbf{return} {true}\;
    \ForEach{$(s, v) \in E \cup E^s$}{
      \eIf{$L(s, v)=\llbracket_i$}{
          \textbf{if} {\texttt{Tabulate}($G$, $v$, $t$)} \textbf{then} \textbf{return} {true}\;
      }{
          \textbf{if} {\texttt{Query}($G$, $v$, $t$)} \textbf{then} \textbf{return} {true}\;
      }
    }
    \textbf{return} {false}\;
  }
  \Proc{\T{$G=(V, E)$, $s\in V$, $t\in V$}}{
    \textbf{if} {$s$ has been visited before by this procedure} \textbf{then} \textbf{return} {false}\;
    \textbf{if} {$s=t$} \textbf{then} \textbf{return} {true}\;
    \ForEach{$(s, v) \in E \cup E^s$}{
      \If{$L(s, v)\ne \enskip\rrbracket_i$}
      {
          \textbf{if} {\texttt{Tabulate}($G$, $v$, $t$)} \textbf{then} \textbf{return} {true}\;
      }
    }
    \textbf{return} {false}\;
  }
\end{algorithm}

\subsection{Indexing Schemes for Conventional Graph Reachability}
\label{subsec:gr}

Quickly answering conventional reachability queries has been the focus of research for over thirty years
due to its wide spectrum of applications.
In order to tell whether a vertex can reach another in a directed graph,
in general, we can use two ``extreme approaches''.
The first approach can answer any query in $O(1)$ time. However, it comes at the cost of quadratic time and space for computing and storing the transitive closure.
The other approach
traverses the graph by depth-first or breadth-first search, attempting to 
find a path between two vertices, which takes linear time and space for each query.
This is apparently very slow for frequent queries on a large graph.

Recent studies on indexing schemes aim to find a promising trade-off lying in-between the two extremes,
reducing the pre-computation time and storage with ``almost'' constant answering time.
We can put the large amount of indexing schemes into two groups: 
\textbf{(1) compression of transitive closure}~(e.g., \cite{wang2006dual,jin20093,jin2011path,cohen2003reachability,chen2008efficient})
and \textbf{(2) pruned search}~(e.g., \cite{yildirim2010grail,chen2005stack,seufert2013ferrari,wei2014reachability}).

\subsubsection{Compression of Transitive Closure}

Approaches in the first group aim to reduce the time and space cost of computing and storing the transitive closure.
For instance,
assuming $k$ is a variable far less than $|V|$,
the dual-labeling method takes $O(|V| + |E| + k^3)$ time to compress the size of transitive closure from $O(|V|^2)$ to $O(|V| + k^2)$,
and preserve the capability of answering each reachability query in constant time~\cite{wang2006dual}.

\begin{figure}
	\centering
	\includegraphics[width=0.7\columnwidth]{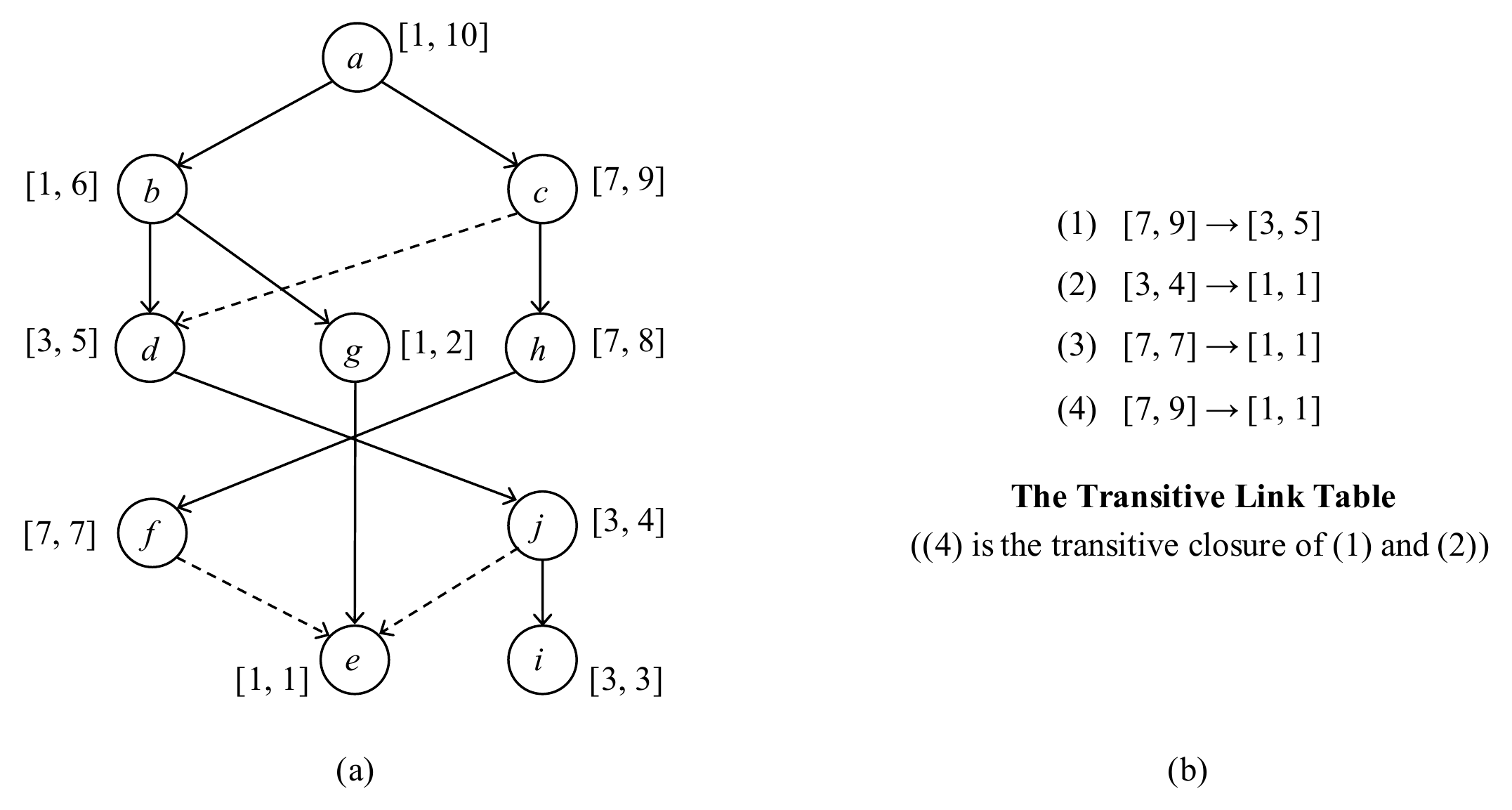}
	\caption{A transitive-closure-compression-based indexing scheme~\cite{wang2006dual}. (a) A graph where one of its spanning trees is represented by the solid edges and non-tree edges are represented by the dashed edges. Vertices are labeled by intervals for reachability queries. (b) The transitive link table for encoding reachability relations implied by the non-tree edges.}
	\label{fig:pre_duallabel}
\end{figure}

\begin{example}\label{ex:dual}
	Figure~\ref{fig:pre_duallabel} illustrates the dual-labeling method, which firstly finds a spanning tree on a given directed acyclic graph\footnote{Since vertices in a strongly connected component (SCC) are reachable from each other, when indexing reachability, we can merge all vertices in every SCC into a single vertex to obtain a directed acyclic graph.}
	and labels each vertex $v$ with an interval $[L_v, H_v]$ according to a traversal on the tree.
	For each interval, $H_v$ is the rank of the vertex $v$ in a post-order traversal of the tree, where the ranks are assumed to begin at 1;
	$L_v$ denotes the lowest rank for any vertex in the sub-tree rooted at $v$.
	This approach guarantees that the vertex $v$ is reachable from the vertex $u$ on the tree if and only if $[L_v, H_v]\subseteq [L_u, H_u]$,
	because the post-order traversal enters a vertex before all its descendants and leaves after visiting all of its descendants.
	
	For each non-tree edge, we record it in a transitive link table as illustrated in Figure~\ref{fig:pre_duallabel}(b),
	and compute a transitive closure. For instance, since $[7, 9]\rightarrow [3, 5]$ and $[3, 4] \rightarrow [1, 1]$ are non-tree edges,
	and $[3, 4]\subseteq [3, 5]$, we also include $[7, 9]\rightarrow [1, 1]$ in the table.
	We then can determine the reachability relation on the graph as follows:
	the vertex $v$ is reachable from the vertex $u$ on the graph if and only if $[L_v, H_v]\subseteq [L_u, H_u]$
	or there exists an entry $[L_x, H_x]\rightarrow [L_y, H_y]$ in the link table
	such that $[L_x, H_x]\subseteq [L_u, H_u] \land [L_v, H_v]\subseteq [L_y, H_y]$.
	
	Assuming the graph has $k$ non-tree edges, we need $O(k^3)$ time to compute the transitive link table of size $O(k^2)$.
	\citet{wang2006dual} showed that, when answering a reachability query, it is not necessary to take $O(k^2)$ time to find the entry $[L_x, H_x]\rightarrow [L_y, H_y]$ in the link table.
	It is actually a special range-temporal aggregation problem and can be solved in $O(1)$ time.
	Thus, we can answer each reachability query in constant time.\hfill $\square$
\end{example}

\subsubsection{Pruned Search}

The pruned-search-based indexing schemes pre-compute information to speed up the depth-first or breadth-first graph traversal by pruning unnecessary searches.
Grail is a typical indexing scheme in this group that can scale to very large graphs~\cite{yildirim2010grail}.
Basically,
it labels each vertex with a constant number of intervals.
We can tell if a vertex is NOT reachable from another by testing the interval containment.
For reachable cases,
it falls back to a graph traversal but is capable of using the intervals to prune unreachable paths.

\begin{figure}
	\centering
	\includegraphics[width=0.7\columnwidth]{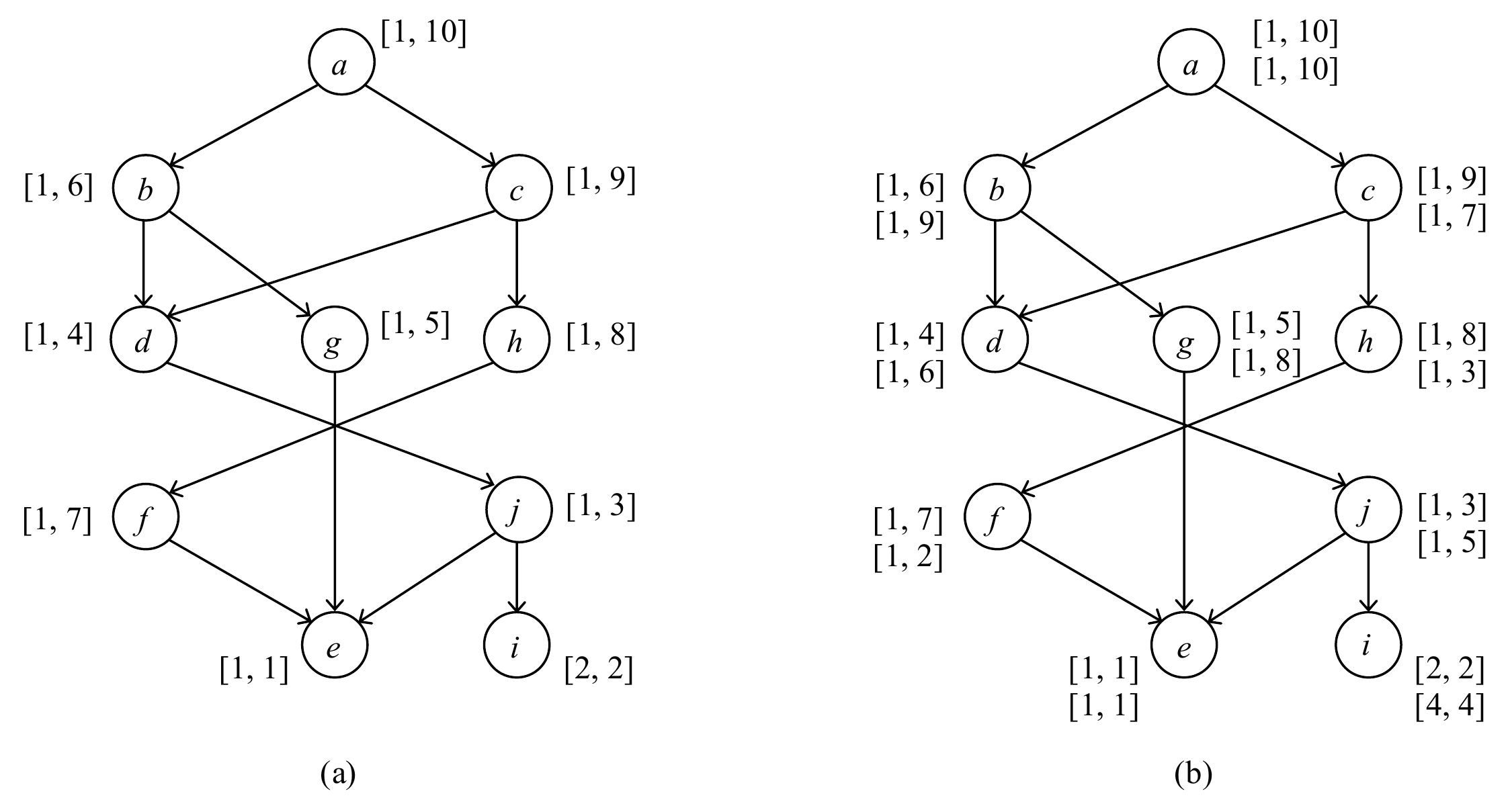}
	\caption{A pruned-search-based indexing scheme~\cite{yildirim2010grail}. (a) A graph where each vertex is labeled by an interval for reachability query. (b) Vertices are labeled by multiple intervals to reduce false positives.}
	\label{fig:pre_grail}
\end{figure}

\begin{example}
	Figure~\ref{fig:pre_grail} illustrates the Grail indexing scheme.
	Given a directed acyclic graph,
	it labels each vertex $v$ with an interval $[L_v, H_v]$ as seen in Example~\ref{ex:dual} but based on a post-order traversal on the graph, as illustrated in Figure~\ref{fig:pre_grail}(a).
	The basic idea is that, on the directed acyclic graph,
	although $[L_v, H_v]\subseteq [L_u, H_u]$ cannot imply that the vertex $v$ is reachable from the vertex $u$,
	$[L_v, H_v]\not\subseteq [L_u, H_u]$ is sufficient to imply that the vertex $v$ is NOT reachable from the vertex $u$.
	For instance, in Figure~\ref{fig:pre_grail}(a),
	$[L_j, H_j]=[1,3]\subseteq [1,8]=[L_h, H_h]$, but the vertex $j$ is not reachable from the vertex $h$.
	
	To prune such false positives implied by the interval containment,
	Grail performs a randomized post-order traversal on the graph multiple times,\footnote{We can randomly order the children of each vertex during the post-order graph traversal.}
	leading to multiple interval labels as shown in Figure~\ref{fig:pre_grail}(b).
	With multiple interval labels,
	we can easily determine that the vertex $j$ is not reachable from the vertex $h$, because the second interval of the vertex $j$, $[1,5]$,
	is not a subset of the second interval of the vertex $h$, $[1,3]$.\hfill $\square$
\end{example}

\subsection{Gaps between the Indexing Schemes and CS-Reachability}
\label{subsec:gaps}

The aforementioned indexing schemes can easily accelerate conventional reachability queries on a common directed graph.
However, owing to the edge labels and the constraint brought by the context-free grammar,
we cannot use them to speed up CS-reachability queries on a program-valid graph, unless we can address the following problem, i.e., reduce the CS-reachability problem to a conventional reachability problem:

\medskip
\noindent
\fbox{
\parbox{0.98\textwidth}{\it
    Problem statement: Given a program-valid graph $G=(V, E)$,
    find a common directed graph $\mathcal{G} = (\mathcal{V}, \mathcal{E})$ and two functions $src: V \mapsto \mathcal{V}$ and $dst: V \mapsto \mathcal{V}$,
    such that, for any pair of vertices, $u, v\in V$ on the program valid graph $G$, the vertex $v$ is CS-reachable from the vertex $u$ if and only if the vertex $dst(v)$ is reachable from the vertex $src(u)$ on the common directed graph $\mathcal{G}$.
}
}
\smallskip

The following sections illustrate an efficient reduction of linear complexity from CS-reachability to conventional graph reachability, which allows us to directly profit from the reachability indexing schemes discussed before.

\section{Overview}
\label{sec:overview}

Figure \ref{fig:overview_ex}(a) shows a program-valid graph with a summary edge from the vertex $b$ to the vertex $d$.
The program-valid graph can be regarded as a data-dependence graph of the code in the figure --- each directed edge represents a data-dependence relation and the parentheses $\llbracket_i$ and $\rrbracket_i$ respectively stand for the call and the return at the $i$th call site.
The summary edge is added over the call site \textit{d = baz(b)}, connecting the input $b$ and the output $d$.
We use the example to illustrate our approach in Section~\ref{subsec:reduction_in_nutshell} and discuss the intuition of its correctness in Section~\ref{subsec:correctness}.

\subsection{Reduction in a Nutshell}
\label{subsec:reduction_in_nutshell}

Our approach, namely \textsc{Flare}, reduces the CS-reachability problem on the program-valid graph to the conventional reachability problem on a common directed graph we refer to as the indexing graph.
This reduction allows us to transform a CS-reachability query to an equivalent conventional reachability query on the indexing graph.
Thus, we then can directly use the indexing schemes introduced in Section~\ref{subsec:gr} for optimization. We focus on addressing two problems: (1) how to build the indexing graph, and (2) how to transform a CS-reachability query to an equivalent query of conventional reachability.

\begin{figure}[t]
	\centering
	\includegraphics[width=0.93\columnwidth]{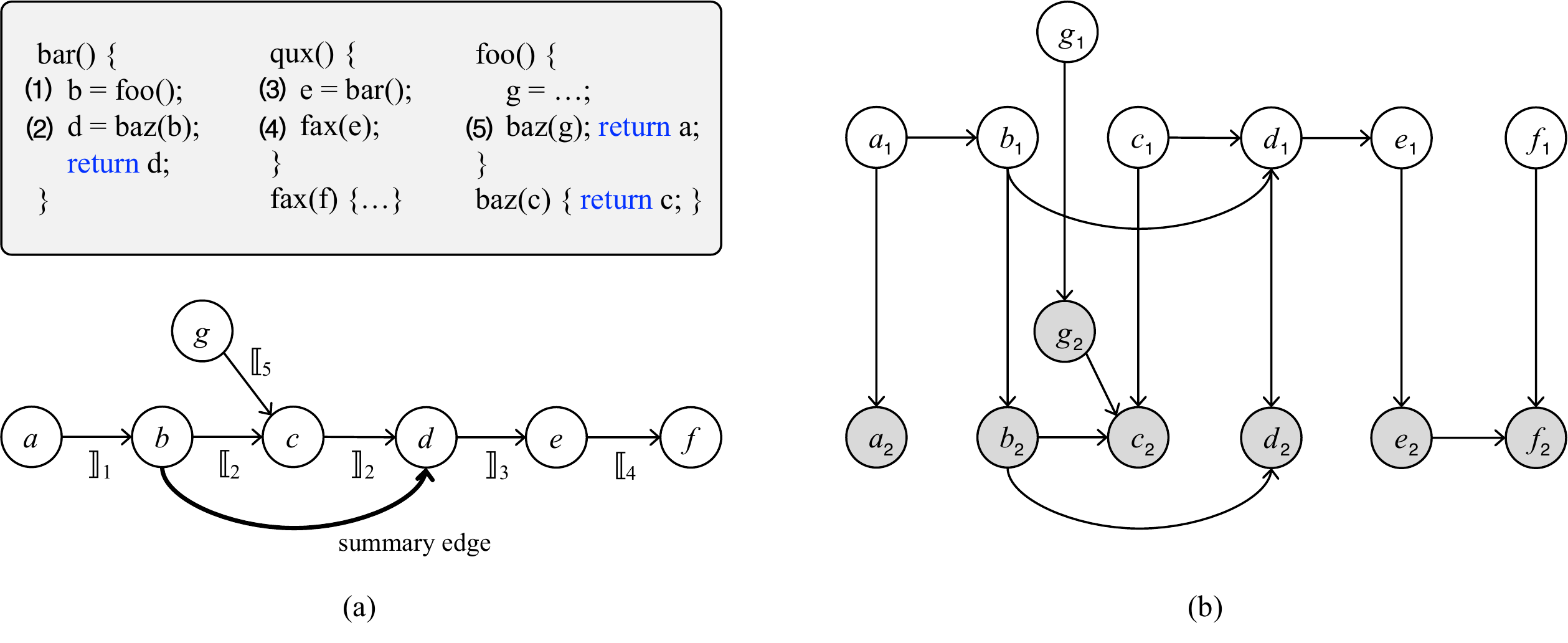}
	\caption{(a) A program-valid graph with one summary edge. The program-valid graph can be regarded as a data-dependence graph of the code shown above --- each directed edge represents a data-dependence relation and the parentheses $\llbracket_i$ and $\rrbracket_i$ respectively stand for the call and the return at the $i$th call site. (b) The indexing graph we build to reduce CS-reachability to conventional graph reachability. Each CS-reachability query on the program-valid graph is equivalent to a conventional reachability query on the indexing graph.}
	\label{fig:overview_ex}
\end{figure}

\smallskip
\textbf{Building the Indexing Graph.}
As shown in Figure \ref{fig:overview_ex}(b),
the indexing graph consists of two copies of the original program-valid graph.
The copies of each vertex $v$ are distinguished by the subscripts $v_1$ and $v_2$.
In the first copy,
we remove all edges labeled by the left-parentheses, which are known as the call edges.
In the second copy,
we remove all edges labeled by the right-parentheses, which are known as the return edges.
For each vertex $v$, we add an edge from the first copy $v_1$ to the second copy $v_2$.
All edge labels are removed from the indexing graph.

\smallskip
\textbf{Querying Context-Sensitive Reachability.}
To answer a CS-reachability query, $\mathcal{Q}(u, v)$, which returns true if and only if the vertex $v$ is CS-reachable from the other vertex $u$ on the program-valid graph, 
we only need to tell if the vertex $u_1$ and the vertex $v_2$ have a conventional reachability relation on the indexing graph.
Let us use the following queries to illustrate the idea.
\begin{enumerate}
	\item \underline{$\mathcal{Q}(a, f) = \textit{{true}}$}. The vertex $f$ is CS-reachable from the vertex $a$
	because there exists a path from the vertex $a_1$ to the vertex $f_2$ on the indexing graph.
	Checking the CS-reachability on the program-valid graph, we can find a $PN$-path $(a,b,c,d,e,f)$ labeled by the string, $\rrbracket_1\llbracket_2\rrbracket_2\rrbracket_3\llbracket_4$,
	that can be derived from the context-free grammar.\smallskip
	
	\item \underline{$\mathcal{Q}(g, c) = \textit{{true}}$}. 
	The vertex $c$ is CS-reachable from the vertex $g$
	because there exists a path from the vertex $g_1$ to the vertex $c_2$ on the indexing graph.
	Checking the CS-reachability on the program-valid graph, we can find an $N$-path $(g,c)$ labeled by the parenthesis, $\llbracket_5$,
	that can be derived from the context-free grammar.
	\smallskip
	
	\item \underline{$\forall v \in V\setminus \{c\}: \mathcal{Q}(g, v) = \textit{{false}}$}. The vertex $g$ cannot context-sensitively reach 
	any vertex except the vertex $c$ on the program-valid graph.
	For example, $\mathcal{Q}(g, f) = \textit{{false}}$ because,
	on the program-valid graph, the path from the vertex $g$ to the vertex $f$ has mismatched parentheses, i.e., $\llbracket_5\rrbracket_2$.
	On the indexing graph,
	there is no path from the vertex $g_1$ to the vertex $f_2$.
\end{enumerate}

\subsection{Intuition of the Correctness}
\label{subsec:correctness}

Using the previous example, this section discusses the intuition of why the reduction is always correct. 
The discussion here serves as a warm-up construction for our formalization in the next section.
Intuitively, the rationale behind our approach is that,
each CS-reachable path (i.e., $P$-path, $N$-path, or $PN$-path) on the program-valid graph corresponds to a path on the indexing graph, and vice versa.
Thus, we can safely transform any CS-reachability query to an equivalent query of conventional graph reachability on the indexing graph.

\begin{figure}[t]
    \centering
    \includegraphics[width=0.93\columnwidth]{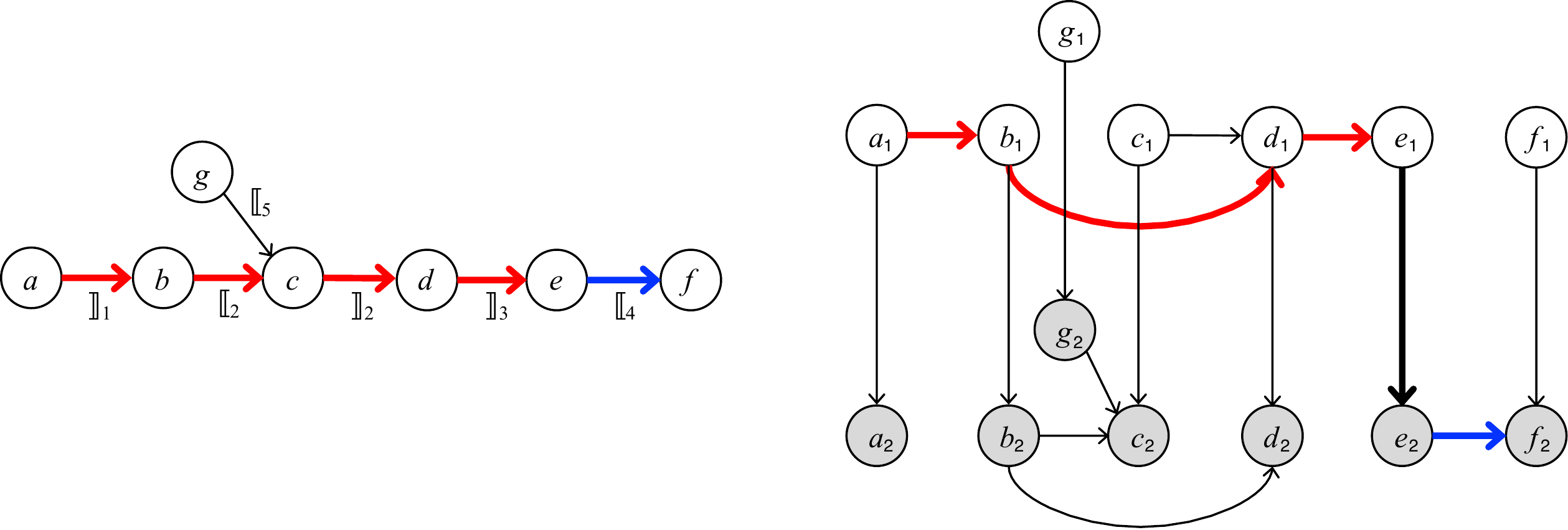}
    \caption{Illustration of the correctness of our approach.
    Mapping the \textbf{\textit{P}-path} $(a, b, c, d, e)$, i.e., the red path, from the program-valid graph to the indexing graph. The sub-path $(b, c, d)$ is replaced with the summary edge $(b_1, d_1)$ on the indexing graph. 
    Mapping the \textbf{\textit{N}-path} $(e, f)$, i.e., the blue path, from the program-valid graph to the indexing graph.
    Mapping the \textbf{\textit{PN}-path}, the concatenation of the $P$-path and the $N$-path, between the program-valid graph and the indexing graph. The $P$-path and the $N$-path are connected via the vertex $e$ on the program-valid graph and via the edge $(e_1, e_2)$ on the indexing graph.}
    \label{fig:overview_correctness}
\end{figure}

Specifically, in the indexing graph,
the first copy of the program-valid graph removes all edges labeled by the left-parentheses so that it contains and only contains $P$-paths.
In other words, each $P$-path on the program-valid graph corresponds to a path on its first copy.
As illustrated in Figure~\ref{fig:overview_correctness},
the $P$-path, $(a, b, c, d, e)$, which is labeled by $\rrbracket_1\llbracket_2\rrbracket_2\rrbracket_3$ on the program-valid graph, corresponds to the path $(a_1, b_1, d_1, e_1)$ on the indexing graph.
Note that the sub-path $(b, c, d)$ is replaced with the summary edge $(b_1, d_1)$ on the indexing graph.

Similarly, in the indexing graph,
the second copy of the program-valid graph removes all edges labeled by the right-parentheses so that it contains and only contains $N$-paths.
In other words, each $N$-path on the program-valid graph corresponds to a path on its second copy.
As illustrated in Figure~\ref{fig:overview_correctness},
the $N$-path, $(e, f)$, which is labeled by $\llbracket_4$ on the program-valid graph, corresponds to the path $(e_2, f_2)$ on the indexing graph.

Other edges from the vertex $v_1$ to the vertex $v_2$ on the indexing graph
connect the $P$-paths and the $N$-paths,
producing the $PN$-paths.
For instance, in Figure~\ref{fig:overview_correctness}, the $PN$-path $(a, b, c, d, e, f)$ on the program-valid graph can be split into two sub-paths,
the $P$-path $(a, b, c, d, e)$ and the $N$-path $(e, f)$,
which respectively
correspond to the path
$(a_1, b_1, d_1, e_1)$ and the path $(e_2, f_2)$ on the indexing graph.

\section{Formalization}
\label{sec:reduction}

In this section, we formally present the idea of building the indexing graph (Section~\ref{subsec:ig}),
as well as how the indexing graph enables the reduction from the CS-reachability problem to the conventional graph reachability problem (Section~\ref{subsec:reduce}).
In the end, we discuss an optimization of the reduction (Section~\ref{subsec:opt}).

\subsection{Indexing Graph}
\label{subsec:ig}

As discussed before,
to reduce CS-reachability to conventional graph reachability,
we need to build the indexing graph based on the program-valid graph.
As a precondition,
we need to compute all summary edges for a given program-valid graph.
\citet{reps1994speeding} have presented an algorithm to compute the summary edges and proved the following lemma about the cost of the algorithm.

\begin{lemma}[Complexity of the Summary Edges~\cite{reps1994speeding}]\label{lem:se}
    The number of summary edges is bounded by $O(\alpha^2|V|)$ and the time to build all summary edges is bounded by $O(\alpha|E| + \alpha^3|V|)$.
\end{lemma}

Since $\alpha$, which stands for the number of function parameters and return values, is a constant in practice,
we can build summary edges efficiently using almost linear time and space with respect to the graph size. We can then define the indexing graph by construction.

\begin{definition}[Indexing Graph]\label{def:ig}
	Given a program-valid graph $G=(V, E)$ where $E=E^\epsilon\cup E^l \cup E^r$ is the union set
	of edges labeled by $\epsilon$, left-parentheses, and right-parentheses,
	and the set $E^s$ of summary edges, 
	an indexing graph
	$\mathcal{G}=(\mathcal{V}, \mathcal{E})$ can be built in the following two steps:
	\begin{enumerate}
		\item Build two copies of the program-valid graph as well as the summary edges:
		\begin{itemize}
			\item $G_1 = (V_1, E_1 = E_1^\epsilon \cup E_1^r \cup E_1^s)$,
			\item $G_2 = (V_2, E_2 = E_2^\epsilon\cup E_2^l\cup E_2^s)$,
		\end{itemize}
		where $V_i$ is a copy of the vertex set $V$, and $E_i^\epsilon$, $E_i^r$, $E_i^l$, and $E_i^s$ are copies of $E^\epsilon$, $E^r$, $E^l$, and $E^s$ over the vertex set $V_i$, respectively.
		
		\item Build the indexing graph $\mathcal{G}=(\mathcal{V}, \mathcal{E})$:
		\begin{itemize}
			\item $\mathcal{V} = V_1\cup V_2$,
			\item $\mathcal{E} = E_1\cup E_2 \cup \{ (v_1, v_2) : v_1\in V_1, v_2\in V_2 \}$,
		\end{itemize}
		 where we use $v_i$ to represent the copy of the vertex $v\in V$ in $V_i$.
	\end{enumerate}
\end{definition}

The following lemma establishes the fact that the size of the indexing graph is linear to the size of the original program-valid graph in practice.

\begin{lemma}[Complexity of the Indexing Graph]
	\label{lemma:linear}
	Assuming $\alpha$ is a constant, we have the space complexity of the indexing graph, $O(|V| + |E|)$, and the time complexity of building the indexing graph, $O(|V| + |E|)$.
\end{lemma}

\begin{proof}
We analyze the complexity of the indexing graph from two aspects, the space complexity and the time complexity.

\smallskip
\textbf{Space Complexity.}
For vertices, the indexing graph contains and only contains two copies of the vertices in the original program-valid graph.
Thus, we have $O(|\mathcal{V}|) = O(|V|)$.
For edges, we have the following equations:

\begin{minipage}[c]{0.7\textwidth}
\begin{equation*}
\begin{split}
O(|\mathcal{E}|)    & = O(|E_1| + |E_2| + |\{ (v_1, v_2) : v_1\in V_1, v_2\in V_2 \}|)  \\
                    & = O((|E| - |E^l| + |E^s|) + (|E| - |E^r| + |E^s|) + |V|)        \\
                    & = O(2|E| + 2|E^s|  + |V|)    \\
                    & = O(2|E| + 2\alpha^2|V| + |V|)  \\
                    & = O(|E| + |V|) \\
\end{split}
\end{equation*}
\end{minipage}
\begin{minipage}[c]{0.3\textwidth}
\begin{equation*}
\begin{split}
\triangleright&~~  \mbox{Definition \ref{def:ig}}\\
\triangleright&~~  \mbox{Definition \ref{def:ig}}     \\
\triangleright&~~  E^r, E^l \subseteq E\\
\triangleright&~~  \mbox{Lemma \ref{lem:se}}\\
\triangleright&~~  \mbox{Constant } \alpha\\
\end{split}
\end{equation*}
\end{minipage}

~\\Putting $O(|\mathcal{V}|)$ and $O(|\mathcal{E}|)$ together, we have $O(|\mathcal{V}| + |\mathcal{E}|) = O(|V| + |E|)$.

\smallskip
\textbf{Time Complexity.} Before building the vertices in $\mathcal{V}$ and the edges in $\mathcal{E}$,
we need to compute the summary edges, of which the time complexity is $O(\alpha|E| + \alpha^3|V|)$ according to Lemma~\ref{lem:se}. Thus the time complexity of building the indexing graph is the sum of $O(\alpha|E| + \alpha^3|V|)$ and $O(|\mathcal{V}| + |\mathcal{E}|)$,
which is $O(|V| + |E|)$ if $\alpha$ is a constant.
\end{proof}

\subsection{Query of CS-Reachability}
\label{subsec:reduce}

The indexing graph allows us to answer CS-reachability queries according to the conventional reachability relations on the indexing graph.
That is,
given a program-valid graph $G=(V, E)$ and its indexing graph $\mathcal{G}=(\mathcal{V}=V_1\cup V_2, \mathcal{E})$,
to determine if 
a vertex $v\in V$ is CS-reachable from a vertex $u\in V$ on the program-valid graph,
we only need to check
if the vertex $v_2\in V_2$ is reachable from the vertex $u_1\in V_1$ on the indexing graph.

To prove the correctness of this claim, we first define two mappings, $\jmath_1$ and $\jmath_2$, as well as their inverse mappings, $\jmath_1^{-1}$ and $\jmath_2^{-1}$.
They provide mappings between the paths (or the edges) on the input program-valid graph and the edges on the indexing graph.
According to Definition~\ref{def:ig}, we can establish the mappings as follows.
$$
\jmath_1(u,\ldots,v) = (u_1, v_1) \hspace{1.25cm} \mbox{\textbf{if} } (u,v)\in E^\epsilon \cup E^r \cup E^s
$$
$$
\jmath_1^{-1}(u_1,v_1) = \left\{
\begin{array}{ll}
(u, v) &  \mbox{ \textbf{if} } (u_1,v_1) \in E_1^\epsilon \cup E_1^r\\
(u,\ldots, v) & \mbox{ \textbf{if} } (u_1,v_1)\in E_1^s
\end{array}
\right.
$$

$$
\jmath_2(u,\ldots,v) = (u_2, v_2) \hspace{1.25cm} \mbox{\textbf{if} } (u,v)\in E^\epsilon \cup E^l \cup E^s
$$
$$
\jmath_2^{-1}(u_2,v_2) = \left\{
\begin{array}{ll}
(u, v) &  \mbox{ \textbf{if} } (u_2,v_2) \in E_2^\epsilon \cup E_2^l\\
(u,\ldots, v) & \mbox{ \textbf{if} } (u_2,v_2)\in E_2^s
\end{array}
\right.
$$

\smallskip
The mapping $\jmath_1$ defined above states that, 
if the path $(u,\ldots,v)$ is a single edge $(u, v)$ labeled by $\epsilon$ or a right-parenthesis,
it is copied to $G_1$ when building the indexing graph.
If $(u,\ldots,v)$ is a summary path,
we create a corresponding summary edge $(u_1, v_1)$ on the copy $G_1$.
The mapping $\jmath_1^{-1}$
states that, for each edge on the copy $G_1$,
if it is a summary edge, it can be mapped back to at least one summary path;
otherwise, it is mapped back to the original edge on the program-valid graph.
Similarly,
we establish the other two mappings, $\jmath_2$ and $\jmath_2^{-1}$, for the copy $G_2$.

\smallskip
Next, we prove the correctness of our CS-reachability query in two steps, i.e., the necessity and the sufficiency.

\begin{lemma}[Necessity]
	\label{lemma:=>}
	Given a program-valid graph $G=(V, E)$ and its indexing graph $\mathcal{G}=(\mathcal{V}=V_1\cup V_2, \mathcal{E})$,
	if there is a CS-reachable path from $u\in V$ to $v\in V$ on the program-valid graph,
	then 
	there must exist
	a path from $u_1\in V_1$ to $v_2\in V_2$ on the indexing graph.
\end{lemma}
\begin{proof}
	According to our discussion before, a CS-reachable path on the program-valid graph may be a $P$-path, $N$-path, or $PN$-path,
	which are discussed below, respectively.
	
	\medskip
	\textbf{Case 1.}
	The production of ``$P\rightarrow MP\enskip |\enskip \rrbracket_iP \enskip|\enskip \epsilon$'' states that,
	any $P$-path $(u, \ldots, v)$ can be partitioned into multiple segments, say 
	$$(u, \ldots, x)(x,\ldots, y)(y,\ldots, z)\ldots(\ldots, v)$$
	where
	each segment is either a summary path or an edge in $E^\epsilon\cup E^r$.
	Thus,
	applying the mapping $\jmath_1$ to each segment will result in a path on $G_1$:
	$$\jmath_1(u, \ldots, x)\jmath_1(x,\ldots, y)\jmath_1(y,\ldots, z)\ldots\jmath_1(\ldots, v) = (u_1,\ldots, v_1).$$
	Since $(v_1, v_2) \in \mathcal{E}$,
	we can find a path $(u_1,\ldots,v_1,v_2)$ on the indexing graph.
	
	\medskip
	\textbf{Case 2.} The production of ``$N\rightarrow MN\enskip |\enskip \llbracket_iN \enskip|\enskip \epsilon$'' states that,
	any $N$-path $(u, \ldots, v)$ can be partitioned into multiple segments, say 
	$$(u, \ldots, x)(x,\ldots, y)(y,\ldots, z)\ldots(\ldots, v)$$
	where
	each segment is either a summary path or an edge in $E^\epsilon\cup E^l$.
	Thus,
	applying the mapping $\jmath_2$ to each segment will result in a path on $G_2$:
	$$\jmath_2(u, \ldots, x)\jmath_2(x,\ldots, y)\jmath_2(y,\ldots, z)\ldots\jmath_2(\ldots, v) = (u_2,\ldots, v_2).$$
	Since $(u_1, u_2) \in \mathcal{E}$,
	we can find a path $(u_1,u_2\ldots,v_2)$ on the indexing graph.
	
	\medskip
	\textbf{Case 3.}
	Given a $PN$-path, $(u, \ldots, v)$,
	according to the context-free grammar,
	we can split it into two segments, say $(u, \ldots, x)(x,\ldots, v)$,
	where the segment $(u, \ldots, x)$ is a $P$-path, and the segment $(x,\ldots, v)$ is an $N$-path.
	According to the above discussions,
	we can find a path $(u_1, \ldots, x_1)$ on $G_1$ and a path $(x_2, \ldots, v_2)$ on $G_2$.
	Since $(x_1, x_2) \in \mathcal{E}$,
	we can find a path $(u_1, \ldots, x_1,x_2,\ldots,v_2)$ on the indexing graph.
\end{proof}

\begin{lemma}[Sufficiency]
	\label{lemma:<=}
	Given a program-valid graph $G=(V, E)$ and its indexing graph $\mathcal{G}=(\mathcal{V}=V_1\cup V_2, \mathcal{E})$,
	if there is a path from $u_1\in V_1$ to $v_2\in V_2$ on the indexing graph,
	there must exist a CS-reachable path from $u\in V$ to $v\in V$ on the program-valid graph.
\end{lemma}
\begin{proof} 
	Given a path $(u_1,\ldots, v_2)$ on the indexing graph, by definition,
	it must be in one of the following three forms:
	(1) $(u_1, \ldots, v_1, v_2)$, (2) $(u_1, u_2 \ldots, v_2)$, or (3) $(u_1, \ldots, x_1)(x_1,x_2)(x_2 \ldots, v_2)$.
	
	\medskip
	\textbf{Case 1.}
	The path is in the form of $(u_1, \ldots, v_1, v_2)$. 
	By definition,
	all vertices from $u_1$ to $v_1$ are in $V_1$, and
	each edge $(x_1, y_1)$ on the path is in $E_1^\epsilon \cup E_1^r \cup E_1^s$.
	Since $\jmath_1^{-1}(x_1, y_1)$ is either a summary path $(x,\ldots,y)$ or an edge $(x, y)\in E^\epsilon\cup E^r$ on the original program-valid graph, 
	applying $\jmath_1^{-1}$ to each edge on the path $(u_1, \ldots, v_1)$ results in a $P$-path $(u, \ldots, v)$ on the program-valid graph.
	
	\medskip
	\textbf{Case 2.} The path is in the form of $(u_1, u_2 \ldots, v_2)$.
	By definition,
	all vertices from $u_2$ to $v_2$ are in $V_2$, and
	each edge $(x_2, y_2)$ on the path is in $E_2^\epsilon \cup E_2^l \cup E_2^s$.
	Since $\jmath_2^{-1}(x_2, y_2)$ is either a summary path $(x,\ldots,y)$ or an edge $(x, y)\in E^\epsilon\cup E^l$ on the original program-valid graph, 
	applying $\jmath_2^{-1}$ to each edge on the path $(u_2, \ldots, v_2)$ results in a $N$-path $(u, \ldots, v)$ on the program-valid graph.
	
	\medskip
	\textbf{Case 3.} The path is in the form of $(u_1, \ldots, x_1)(x_1,x_2)(x_2 \ldots, v_2)$.
	Based on the discussion of Case 1 and Case 2,
	the prefix $(u_1, \ldots, x_1)$ corresponds to a $P$-path $(u,\ldots,x)$ on the program-valid graph;
	the suffix $(x_2, \ldots, v_2)$ corresponds to an $N$-path $(x,\ldots,v)$ on the program-valid graph.
	Thus, the concatenation of the two paths, i.e., $(u,\ldots,x,\ldots,v)$, is a $PN$-path on the program-valid graph.
\end{proof}

Putting Lemma~\ref{lemma:linear}, Lemma~\ref{lemma:=>}, and Lemma~\ref{lemma:<=}
together, we have the following theorem that summarizes our result.

\begin{theorem}
	The CS-reachability problem on a program-valid graph can be reduced to a conventional graph reachability problem
	on the indexing graph in linear time and space with respect to the size of the input program-valid graph.
\end{theorem}

\subsection{Saving the Copies of the Program-Valid Graph}
\label{subsec:opt}

Instead of proposing a sophisticated CFL-reachability algorithm like many previous works,
we have presented an approach that simply copies the input program-valid graph twice to build the indexing graph for addressing the CS-reachability problem.
In practice, we can take a further step to make our approach simpler
---
we do not need to physically copy the program-valid graph for building the indexing graph.

Our key insight is that the indexing graph shares the vertices and the edges with the summary-edge-augmented program-valid graph $G=(V, E^\epsilon\cup E^r \cup E^l \cup E^s)$.
Thus, we do not need to physically generate the copies, $G_1$ and $G_2$,
but reuse the data structure of $G$ and
logically distinguish the copies using an extra integer in $\{1, 2\}$.
Algorithm~\ref{alg:ig-v} and Algorithm~\ref{alg:ig-e}
demonstrate our idea of implementing the basic operations over the indexing graph, i.e.,
iterating the vertices and iterating the successors of a given vertex.
The algorithms do not use the physically copied vertices $v_i\in \mathcal{V}$ but represent the vertex
as a pair $(v\in V, i\in \{1, 2\})$, which reuses the vertex $v$ in the program-valid graph.
Similarly,
we can replace a physically copied edge $(v_i, u_j)\in\mathcal{E}$ with a pair $((v, i), (u, j))$.  
Note that Algorithm~\ref{alg:ig-v} and Algorithm~\ref{alg:ig-e} are sufficient for implementing all graph algorithms over the indexing graph including the indexing algorithms discussed in Section~\ref{subsec:gr}. This is
because they have essentially represented the indexing graph as an adjacent list, one primary data structure for graphs.

\begin{algorithm}[t]\small
  \caption{Iterating the vertices of an indexing graph.}
  \label{alg:ig-v}
  \SetKwFunction{Q}{vertices}
  \SetKwProg{Proc}{Procedure}{}{}
  \Proc{\Q{$G=(V, E^\epsilon\cup E^r \cup E^l \cup E^s)$}}{
    \ForEach{$v \in V$}{
      \ForEach{$i \in \{ 1, 2 \}$}{
          do some operation on $(v, i)$; /* iterate all $v_i\in\mathcal{V}$ */\\
      }
    }
  }
\end{algorithm}

\begin{algorithm}[t]\small
  \caption{Iterating the successors of a given vertex on the indexing graph.}
  \label{alg:ig-e}
  \SetKwFunction{Q}{successors}
  \SetKwProg{Proc}{Procedure}{}{}
  \Proc{\Q{$G=(V, E^\epsilon\cup E^r \cup E^l \cup E^s), (v\in V, i\in \{ 1, 2 \})$}}{
    \eIf{$i = 1$}{
        do some operation on $(v, 2)$; /* $\forall v\in V, (v_1, v_2)\in \mathcal{E}$ */\\
        \ForEach{$(v, u)\in E^\epsilon\cup E^r \cup E^s$}{
            do some operation on $(u, 1)$ /* $(v_1, u_1)\in \mathcal{E}$ */\\
        }
    }{
      \ForEach{$(v, u)\in E^\epsilon\cup E^l \cup E^s$}{
            do some operation on $(u, 2)$ /* $(v_2, u_2)\in \mathcal{E}$ */\\
      }
    }
  }
\end{algorithm}

To conclude, in practice, 
the only overhead of building the indexing graph and reducing CS-reachability to conventional graph reachability is to compute the summary edges,
which is a well-studied problem and can be addressed efficiently as stated in Lemma~\ref{lem:se}.
Despite its simplicity,
we show how wide its applicability is in the next section.

\section{Applications}
\label{sec:app}

Our results can speed up a wide range of context-sensitive data flow analyses working on different program-valid graphs.
To show the practicality,
we apply our results to two program analyses, i.e., \citet{lerch2014flowtwist}'s information-flow analysis and \citet{li2013precise}'s alias analysis, as summarized in Table~\ref{tab:app}.
The two analyses work on two different program-valid graphs, which are known as the \textit{exploded super-graph} and the \textit{value-flow graph}, respectively.\footnote{We provide examples to illustrate their program-valid graphs in Appendix~\ref{app:esg} and Appendix~\ref{app:vfg}, respectively.}
Despite many differences,
both applications formulate their problems as CS-reachability queries
and their core engines follow the same spirit of \citet{reps1994speeding,reps1995precise}'s tabulation algorithm, which, as shown in Algorithm~\ref{alg:tabulation}, traverses the underlying program-valid graph for answering CS-reachability queries.
Since each query needs a graph traversal, it is inefficient when CS-reachability is frequently queried.

To improve the query performance, we build the indexing graphs based on their input program-valid graphs
and
employ existing indexing schemes for acceleration.
Given that there are a large number of indexing schemes we can choose as discussed in Section~\ref{subsec:gr},
we summarize the criteria of selecting indexing schemes in the rules below.
Table~\ref{tab:app} summarizes the indexing schemes chosen for the two applications.

\begin{principle}\label{p:grail}
	If a program analysis needs to query both the CS-reachability relation and
	the paths between two vertices,
	we prefer to use the pruned-search-based indexing schemes, which can return paths when responding to CS-reachability queries.
\end{principle}

\begin{principle}\label{p:pt}
	If a program analysis does not need to return any paths when responding to a CS-reachability query,
	we prefer to use the indexing schemes that compress the transitive closure, which often exhibits better query performance than the pruned-search-based indexing schemes.
\end{principle}

\begin{table}[t]
	\footnotesize
	\centering
	\caption{The basics of the two applications and the indexing schemes for them.}\label{tab:app}
	\begin{tabular}{r|c|c}
		\toprule
		\multirow{2}*{\textbf{Application}} & Information-Flow Analysis$^1$ & Alias Analysis$^2$\\
		 & \cite{lerch2014flowtwist} & \cite{li2013precise} \\
		\midrule
		\textbf{Program-Valid Graph} & Exploded Super-graph& Value-Flow Graph \\
		\midrule
		\textbf{Basic Approach} & \multicolumn{2}{c}{\citet{reps1994speeding, reps1995precise}'s Tabulation Algorithm}\\
		\midrule\midrule
		\textbf{Indexing Scheme} &Grail~\cite{yildirim2010grail}&PathTree~\cite{jin2011path} \\
		\midrule
		\textbf{Index Size} &$O(k|V|)$&$O(k|V|)$ \\
		\midrule
		\textbf{Indexing Time} & $O(k(|V| + |E|))$&$O(k|E|)$\\
		\midrule
		\textbf{Query Time} & $O(k)$ or $O(k(|V| + |E|))$&$O(1)$ or $O(\log^2k)$\\
		\bottomrule
				\multicolumn{3}{l}{$^1$ $k\le 5$ denotes how many times we randomly traverse the graph to build the index.}\\
				\multicolumn{3}{l}{$^2$ $k$ is the number of paths that can cover the input graph. }
	\end{tabular}
\end{table}

\textbf{Indexing Context-Sensitive Information-Flow Analysis.}
Information flow is the transfer of information from a variable $x$ to a variable $y$ in a given program, which, in this application, is formulated as a CS-reachability problem over the exploded super-graph.
When answering a CS-reachability query,
this application needs to find one or multiple CS-reachable paths that lead to the information flow.
This is critical when the information-flow analysis is used to 
detect security violations
as we need to check the violation-triggering paths so as to fix the violations.
Therefore,
we follow Rule~\ref{p:grail} to use the pruned-search-based indexing schemes
as they can provide paths as the evidence of reachability.
In the implementation,
we use the Grail indexing scheme~\cite{yildirim2010grail},
which, as illustrated in Example 2.5, builds the reachability index by randomly traversing an input graph $k$ times ($k\le 5$ in practice and we use $k=5$ in the implementation).
The Grail indexing scheme allows us to answer most unreachable queries in $O(k)$ time
and, for reachable queries, due to the pruned search,
we can return a path within linear time and space with respect to the path length.
Apparently,
the reachability indexing scheme significantly improves the query efficiency 
over a normal tabulation algorithm.

\textbf{Indexing Context-Sensitive Alias Analysis.}
Alias analysis statically determines if two pointer variables can point to the same memory location during program execution.
In this application, the aliasing problem is formulated as a CS-reachability problem over the value-flow graph.
Since we often only need to answer ``yes'' or ``no'' when querying if a pointer is the alias of the other,
it is not necessary to find any CS-reachable path between two pointer variables.
Thus,
we follow Rule~\ref{p:pt} to use the indexing schemes that compress the transitive closure.
Specifically,
we use the PathTree indexing scheme,
which is based on a path-decomposition that partitions the input directed graph into $k$ paths~\cite{jin2011path}.
Basically,
the Path-Tree method takes $O(k|E|)$ time to compress the size of transitive closure from $O(|V|^2)$ to $O(k|V|)$.
The compressed transitive closure, a.k.a., the index,
allows us to answer reachability queries in $O(1)$ time for most cases and in $O(\log^2k)$ time for the others.
In the implementation, we use SCARAB~\cite{jin2012scarab}, a unified reachability computation framework, to improve the performance of PathTree. 
Basically, SCARAB improves the performance by reducing the graph size, or more specifically, by extracting a ``reachability backbone'' that carries the major reachability information.
Armed with the indexing scheme, the alias analysis can answer aliasing queries far more efficiently than \citet{li2013precise}'s original approach, 
with just a moderate time and space overhead for building and storing the index.

\section{Evaluation}
\label{sec:eval}

We implemented the context-sensitive information-flow analysis~\cite{lerch2014flowtwist} and context-sensitive alias analysis~\cite{li2013precise},
as well as their indexed counterparts, namely \textsc{FlareIFA} and \textsc{FlareAA},
on top of the LLVM compiler infrastructure~\cite{lattner2004llvm}.
Given an input program,
we compile it to the LLVM bitcode and
follow their original method to build the program-valid graphs, i.e.,
the exploded super-graph and the value-flow graph, respectively.
We then follow \citet{reps1994speeding}'s approach to compute all summary edges and build the indexing graph according to Definition~\ref{def:ig}.
As discussed before,
we use Grail~\cite{yildirim2010grail} and PathTree~\cite{jin2011path} as the indexing schemes for information-flow analysis and alias analysis, respectively.
The code of Grail and PathTree is open source.\footnote{Grail: \url{https://github.com/zakimjz/grail}; PathTree: \url{http://www.cs.kent.edu/~nruan/soft.html}.}
We use their source code in our implementation to avoid unnecessary biases introduced by engineering issues.
The whole artifact for evaluation is publicly available online.\footnote{Artifact for evaluation: \url{https://github.com/qingkaishi/context-sensitive-reachability}.}

\subsection{Experiment Setup}

To demonstrate how our approach improves the performance of the context-sensitive information-flow analysis and the context-sensitive alias analysis,
we conducted a series of experiments over standard benchmark programs and open-source software.

\smallskip
\textbf{Benchmark Programs.}
Our experiments are performed over the programs from SPEC CINT2000, a standard benchmark suite widely used in literature~\cite{henning2000spec} as well as four real-world and much larger programs: \textit{git}, \textit{vim}, \textit{icu}, and \textit{ffmpeg}.\footnote{Git: \url{https://git-scm.com/}; Vim: \url{https://www.vim.org/}; ICU: \url{http://site.icu-project.org/}; FFmpeg: \url{http://ffmpeg.org/}.}
The basic information of the sixteen benchmark programs is listed in Table~\ref{tab:benchmarks},
including the lines of code (LoC) of each program as well as the number of vertices and edges on their program-valid graphs.
As shown in the table, the sizes of the programs range from a few thousand lines of code to nearly one million and a program-valid graph may contain tens of millions of vertices and edges,
which makes it challenging to answer reachability queries quickly, let alone CS-reachability queries.

\smallskip
\textbf{Baseline Approaches.}
For each of the information-flow analysis and the alias analysis,
we organize the experiments in three parts.
In the first two parts,
we show that it is not possible to compute a full transitive closure for answering the CS-reachability queries while we can compute the reachability indexes within a reasonable time and space overhead.
This is because computing a transitive closure requires sub-cubic complexity for general CFL-reachability while we can often build reachability indexes within almost linear time complexity.
The experiments of computing the transitive closure are run with a limit of six hours.
In the third part,
we show that the indexed analyses, i.e., \textsc{FlareIFA} and \textsc{FlareAA},
are much faster than their original counter-parts~\cite{li2013precise,lerch2014flowtwist}.
As discussed before, the original analyses follow the same spirit of \citet{reps1994speeding,reps1995precise}'s tabulation algorithm (see Algorithm~\ref{alg:tabulation}), which, essentially, traverses the input program-valid graph to answer each CS-reachability query.

We notice that there have been a few optimized CFL-reachability algorithms proposed in recent years, particularly for the standard Dyck-CFL-reachability problem~\cite{zhang2013fast,chatterjee2017optimal,yuan2009efficient}. 
However, due to the differences in the underlying CFL and graph structures, their approaches cannot be directly employed in context-sensitive data flow analysis. Thus, we cannot compare our approach to them. Instead, we discuss them in Section~\ref{sec:relatedwork}.

We also notice that there are many query caching mechanisms~\cite{zhou2018cgraph} and
graph simplification algorithms, such as eliminating reachability-irrelevant vertices and edges~\cite{li2020fast}, which can also improve the query performance. It is noteworthy that all these optimizations are orthogonal to our approach. Thus, it does not make any sense for comparison. In practice, our approach can be used together with them for better performance. For instance, our reduction can be performed on a simplified program-valid graph, which will lead to a smaller indexing graph and, thus, faster query speed.

\smallskip
\textbf{Environment.}
All experiments were run on a server with 
eighty ``Intel Xeon CPU E5-2698 v4 @ 2.20GHz'' processors
and 256GB of memory running Ubuntu-16.04.

\begin{table}[t]
	\centering
	\footnotesize
	\caption{Benchmark programs from SPEC CINT 2000.}
	\label{tab:benchmarks}
	\begin{tabular}{c|lr|rr|rr}
		\toprule
		\multirow{2}{*}{\textbf{ID}} & 
		\multirow{2}{*}{\textbf{Program}} & 
		\multirow{2}{*}{\textbf{LoC}} &
		\multicolumn{2}{c|}{Information-Flow Analysis} &
		\multicolumn{2}{c}{Alias Analysis} \\
		&&&
		\textbf{\# Vertices} & \textbf{\# Edges} &
		\textbf{\# Vertices} & \textbf{\# Edges}\\
		\midrule
		1 & mcf      & 2K            & 33.3K   & 35.0K   &22.2K     & 29.4K      \\
		2 & bzip2    & 3K            & 257.3K  & 277.8K  &59.5K     & 76.2K      \\
		3 & gzip     & 6K            & 314.5K  & 332.2K  &135.6K    & 182.8K     \\
		4 & parser   & 8K            & 540.5K  & 566.7K  &574.2K    & 749.1K     \\
		5 & vpr      & 11K           & 3.0M    & 3.4M    &347.7K    & 421.7K     \\
		6 & crafty   & 13K           & 651.0K  & 678.5K  &280.2K    & 362.1K     \\
		7 & twolf    & 18K           & 635.4K  & 696.6K  &468.0K    & 624.0K     \\
		8 & eon      & 22K           & 798.8K  & 969.9K  &766.0K    & 852.0K     \\
		9 & gap      & 36K           & 689.8K  & 788.3K  &3.2M      & 3.8M       \\
		10 & vortex  & 49K           & 659.1K  & 714.2K  &4.4M      & 5.6M       \\
		11 & perlbmk & 73K           & 2.4M    & 2.6M    &9.2M      & 11.7M      \\
		12 & gcc     & 135K          & 9.7M    & 10.7M   &17.0M     & 22.2M      \\
		\midrule
		13 & git-2.32.0     & 248K   & 5.4M    & 5.7M    & 20.6M & 25.4M\\
		14 & vim-8.2.3047   & 386K   & 14.7M   & 19.0M   & 40.0M & 50.0M\\
		15 & icu-69.1       & 594K   & 5.9M    & 6.5M    & 14.4M & 18.2M\\
		16 & ffmpeg-3.0     & 940K   & 5.3M    & 6.2M    & 33.8M & 44.6M\\
		\bottomrule
	\end{tabular}
\end{table}

\subsection{Information-Flow Analysis}

The evaluation of the information-flow analysis is in three parts,
which aim to show that (1)~computing the transitive closure is not practical (Section~\ref{ifa1}), (2) compared to computing a transitive closure, the overhead of computing the reachability index is reasonable (Section~\ref{ifa2}), and (3) the reachability index significantly speeds up CS-reachability queries~(Section~\ref{ifa3}).

\subsubsection{Transitive Closure is not Practical.}\label{ifa1}
For information-flow analysis,
computing a transitive closure allows us to answer an unreachable CS-reachability query (or, in this application, an information-flow query) in constant time.
For reachable cases,
the transitive closure allows us to find paths from a source vertex to a target vertex within linear time and space with respect to the path size.
This is because when searching a path from the source vertex, we can always prune unreachable paths based on the transitive closure.
However,
due to the high complexity of computing the transitive closure~\cite{chaudhuri2008subcubic},
we failed to compute the transitive closure for ten of our sixteen benchmark programs.
Thus, while computing a transitive closure is an ideal approach to the information-flow analysis, it is not practical for analyzing large-scale software.

\subsubsection{Overhead of Indexing is Reasonable.}\label{ifa2}
\begin{figure}[t]
    \centering
    \includegraphics[width=\textwidth]{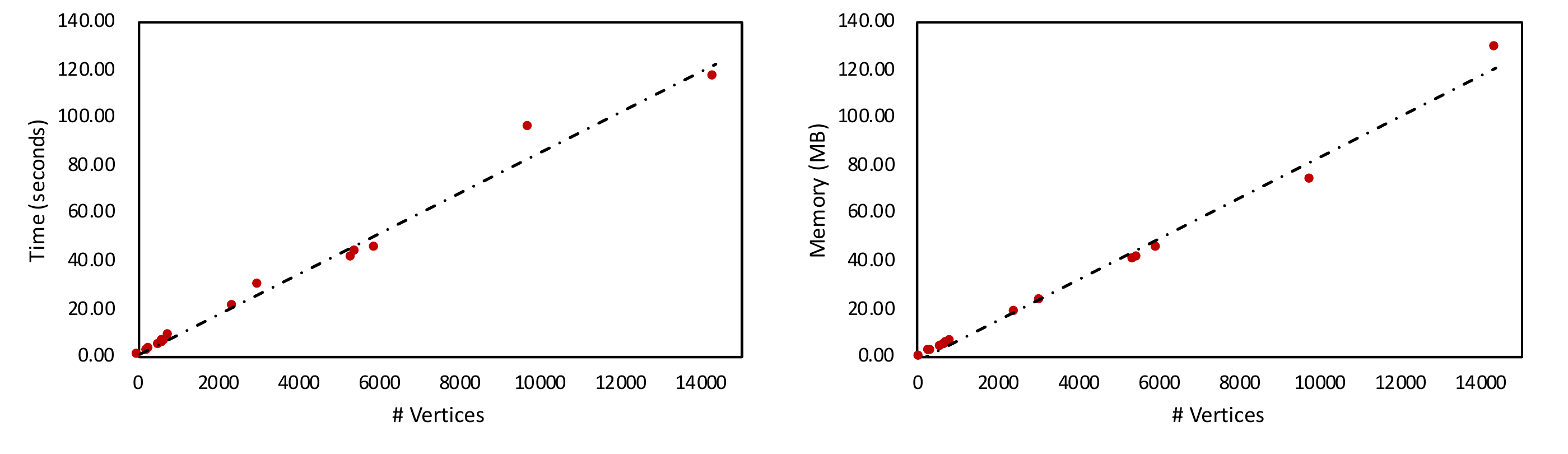}
    \caption{Building the indexing graph exhibits linear complexity in the information-flow analysis.}
    \label{fig:flareifa-ig}
\end{figure}

The indexing procedure for the information-flow analysis includes two parts.
The first part is to compute the indexing graph and the second part is to build the Grail indexes~\cite{yildirim2010grail} on the indexing graph.
As discussed in Section~\ref{subsec:opt}, the main cost of building an indexing graph is to compute the summary edges. 
Table~\ref{tab:flareifa} lists the time and the memory cost of computing the indexing graphs
as well as the cost of computing the Grail indexes.
Figure~\ref{fig:flareifa-ig} shows that
the cost of building an indexing graph is of linear complexity with respect to the input graph size.
For the largest graph that contains tens of millions of vertices, we only need less than 2 minutes and about 80 MB of space to build the indexing graph.

\begin{table}[t]
    \centering\footnotesize
    \caption{The time cost (seconds) and the memory cost (MB) of indexing the information-flow analysis.}
    \begin{tabular}{c|c|c|c|c|c|c}
    \toprule
         \multirow{2}{*}{\textbf{ID}} & \multicolumn{3}{c|}{\textbf{Time}} & \multicolumn{3}{c}{\textbf{Memory}} \\\cline{2-7}
            & Indexing Graph & Grail & \textbf{Total}   &  Indexing Graph  &  Grail  & \textbf{Total}   \\\midrule
1&0.36 &0.03&0.39 &0.25 &1.52 &1.78 \\
2&2.34 &0.33&2.67 &1.96 &11.78 &13.74 \\
3&3.36 &0.24&3.60 &2.40 &14.39 &16.79 \\
4&4.83 &0.42&5.25 &4.12 &24.74 &28.87 \\
5&30.39 &5.25&35.64 &23.20 &139.18 &162.38 \\
6&6.06 &1.26&7.32 &4.97 &29.80 &34.77 \\
7&6.01 &2.49&8.50 &4.85 &29.08 &33.93 \\
8&8.84 &1.56&10.40 &6.09 &36.57 &42.66 \\
9&6.63 &0.78&7.41 &5.26 &31.57 &36.84 \\
10&5.58 &2.16&7.74 &5.03 &30.17 &35.20 \\
11&21.32 &6.3&27.62 &18.38 &110.29 &128.67 \\ 
12&95.92 &22.71&118.63 &74.36 &446.14 &520.50 \\\midrule 
13&43.95 &3.9&47.85 &41.46 &248.76 &290.22 \\
14&117.18 &80.7&197.88 &129.51 &657.08 &786.60 \\
15&45.81 &2.76&48.57 &45.35 &272.12 &317.47 \\
16&41.45 &3.99&45.44 &40.63 &243.80 &284.44 \\
        \bottomrule
    \end{tabular}
    \label{tab:flareifa}
\end{table}

\begin{figure}[t]
    \centering
    \includegraphics[width=\textwidth]{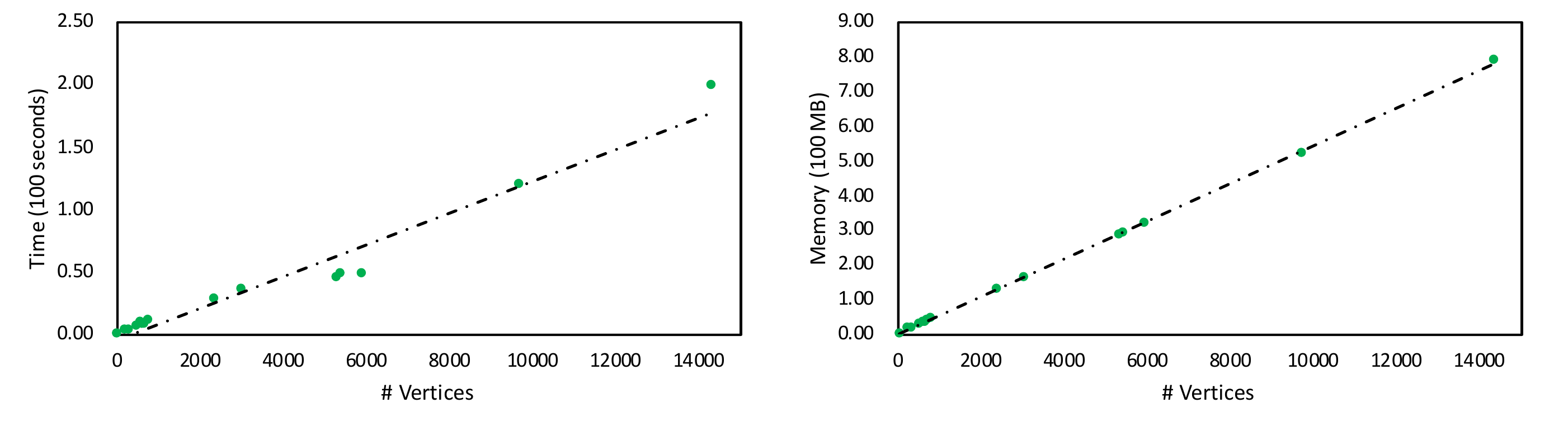}
    \caption{Indexing the information-flow analysis exhibits linear complexity in practice.}
    \label{fig:flareifa-total}
\end{figure}

\begin{figure}[t]
    \centering
    \includegraphics[width=\textwidth]{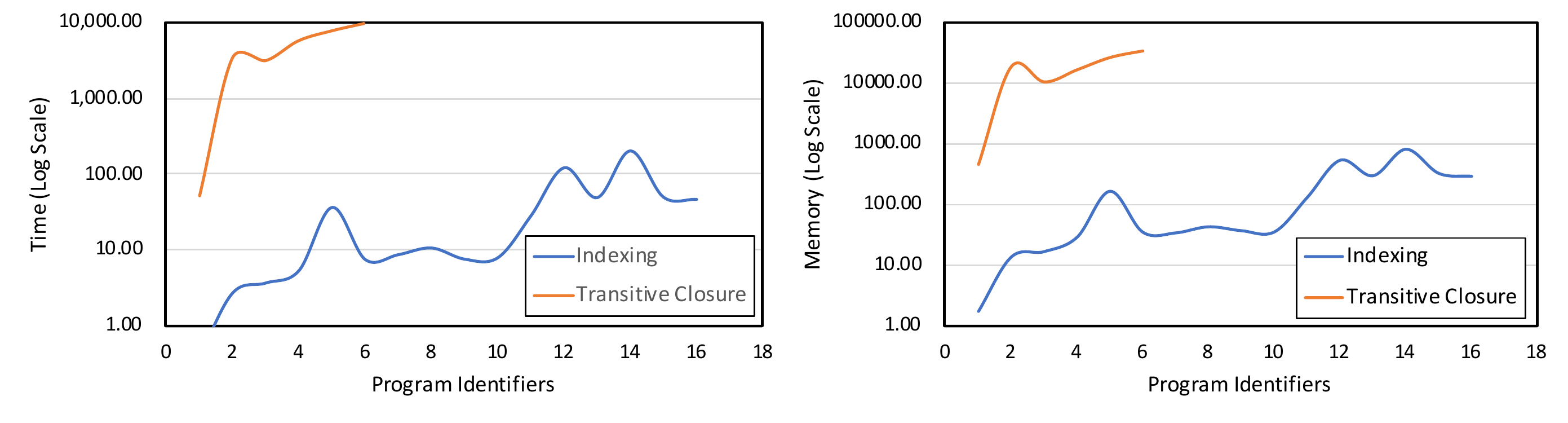}
    \caption{Compared to computing the transitive closure, the overhead of indexing is reasonable for the information-flow analysis.}
    \label{fig:flareifa-tc}
\end{figure}

Figure~\ref{fig:flareifa-total} shows the total space and time we need to build the Grail index.
The space overhead is moderate, as the index size is only about 650MB for our largest program.
Meanwhile,
as illustrated in Figure~\ref{fig:flareifa-total},
the time cost of building the index is of linear complexity with respect to the graph size.
Therefore,
the indexing scheme for the information-flow analysis scales quite gracefully in practice.
Figure~\ref{fig:flareifa-tc} shows that,
in comparison to computing a transitive closure, the overhead of computing the reachability index is negligible in practice.
Even for the first six small programs for which we succeed in computing the transitive closure,
computing the index is 540$\times$ faster and saves 99.7\% of the space compared to computing the transitive closure.

\subsubsection{Indexing Enables Much Faster Queries.}\label{ifa3}
As listed in Table~\ref{tab:flareifa-query}, the reachability index allows 
the indexed information-flow analysis, namely \textsc{FlareIFA},
to answer 10,000 reachable queries in 6,500 milliseconds and 10,000 unreachable queries in 350 milliseconds,
110$\times$ to 7,642$\times$ and 565$\times$ to 36,541$\times$ faster than the baseline approach~\cite{lerch2014flowtwist}.
Readers may notice that answering a reachable query takes much longer than answering an unreachable query.
This is because, for a reachable query in the information-flow analysis,
we need to additionally compute a path from the source vertex to the target vertex.

\begin{table}[t]
	\centering\footnotesize
	\caption{Time cost (milliseconds) of answering 10,000 reachable queries ($R$), 10,000 unreachable queries ($\neg{R}$), and all queries (Total) in the information-flow analysis. We also compute the minimum (min), median (med), and maximum (max) speedup compared to the baseline approach.}
	\label{tab:flareifa-query}
	\begin{tabular}{c|r|r|r|r|r|r}
		\toprule
		\multirow{2}{*}{\textbf{ID}} & \multicolumn{3}{c|}{\textbf{\textsc{FlareIFA}}} & \multicolumn{3}{c}{\cite{lerch2014flowtwist}} \\\cline{2-7}
		   & $R$ & $\neg{R}$ & \textbf{Total} & $R$ & $\neg{R}$ & \textbf{Total}\\
		\midrule
1 &22     &2    &23        &35,505      &2,103      &37,608 \\
2 &150    &62   &212       &34,000      &35,159     &69,159 \\
3 &214    &240  &454       &329,274     &138,917    &468,191 \\
4 &352    &132  &485       &714,638     &150,179    &864,818 \\
5 &2,317  &320  &2,637     &2,755,797   &983,474    &3,739,271 \\
6 &838    &23   &861       &786,158     &41,560     &827,718 \\
7 &290    &9    &299       &31,943      &29,668     &13,611 \\
8 &2,123  &181  &2,304     &558,344     &255,466    &813,810 \\
9 &1,321  &114  &1,435     &1,499,938   &269,286    &1,769,224 \\
10 &1,194 &101  &1,296     &1,729,948   &351,161    &2,081,109 \\
11 &3,654 &205  &3,859     &5,712,390   &193,101    &5,905,491 \\
12 &6,416 &315  &6,731     &21,600,091  &11,495,662 &33,095,753 \\\midrule
13 &1,012 &166  &1,179     &1,504,814   &1,439,299  &2,944,113 \\
14 &2,787 &50   &2,837     &1,824,255   &1,717,196  &3,541,450 \\
15 &1,898 &242  &2,140     &14,504,613  &555,593    &15,060,206 \\
16 &1,532 &213  &1,745     &1,138,623   &1,200,232  &2,338,855 \\
		\midrule\midrule
		\textbf{min} & 110$\times$ &  565$\times$   &206$\times$ & \multicolumn{3}{c}{}\\
		\textbf{med} &  1,319$\times$ & 2,323$\times$  &1,379$\times$ &\multicolumn{3}{c}{N/A}\\
		\textbf{max} &  7,642$\times$ &  36,541$\times$   &7,036$\times$ &\multicolumn{3}{c}{}\\
		\bottomrule
	\end{tabular}
\end{table}

\subsection{Alias Analysis}

Same as the information-flow analysis, the evaluation of the alias analysis also consists of three parts,
aiming to show that (1) computing a transitive closure is not practical (Section~\ref{aa1}), (2) the overhead of computing the reachability index is reasonable (Section~\ref{aa2}), and (3) the reachability index significantly speeds up the CS-reachability queries (Section~\ref{aa3}).

\subsubsection{Transitive Closure is not Practical.}\label{aa1}
Computing a transitive closure allows us to answer both reachable and unreachable CS-reachability queries (or, in this application, aliasing queries) in constant time. However, it is of sub-cubic time complexity and quadratic space complexity to compute a transitive closure~\cite{chaudhuri2008subcubic}, which is unaffordable in practice.
In the experiment, we
finished the computation only for the programs with less than 40 KLoC and
failed for all other larger programs. This fact shows that computing the transitive closure is not practical for alias analysis, either.

\subsubsection{Overhead of Indexing is Reasonable.}\label{aa2}
\begin{figure}[t]
    \centering
    \includegraphics[width=\textwidth]{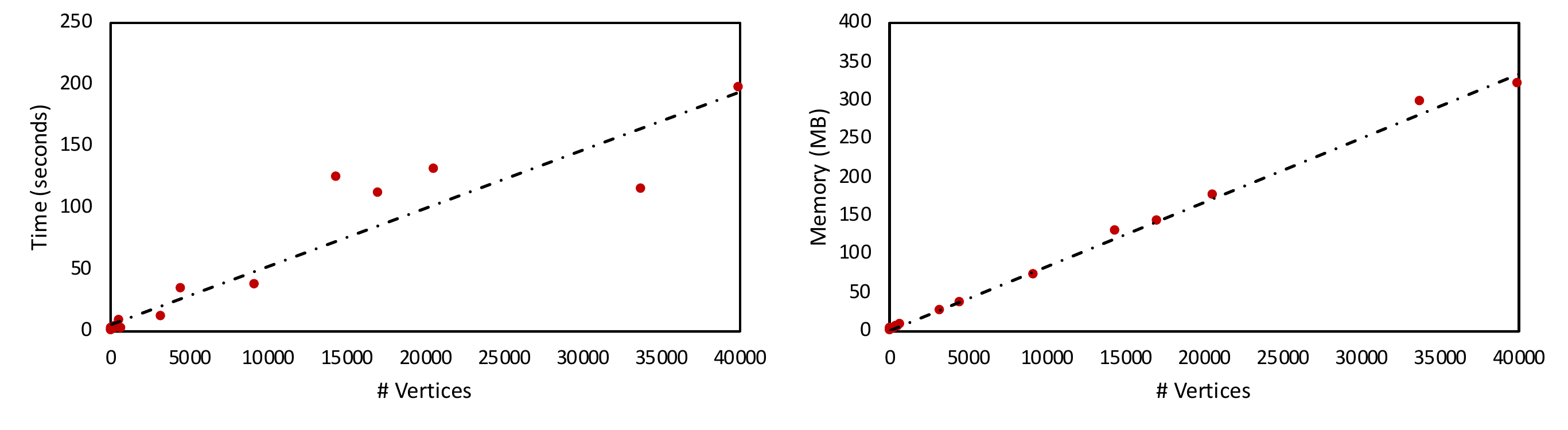}
    \caption{Building the indexing graph exhibits linear complexity in the alias analysis.}
    \label{fig:flareaa-ig}
\end{figure}

Same as the information-flow analysis,
the indexing procedure for the alias analysis also includes two parts.
The first part is to compute the indexing graph and the second part is to employ the PathTree indexing scheme~\cite{jin2011path} over the indexing graph.
As discussed in Section~\ref{subsec:opt}, the main cost of building the indexing graph is to compute the summary edges. 
Table~\ref{tab:flareaa} lists the time and the memory cost of computing the indexing graphs
as well as the cost of computing the PathTree indexes.
As illustrated in Figure~\ref{fig:flareaa-ig},
for alias analysis,
the cost of building the indexing graph is also of linear complexity with respect to the input graph size.
For the largest graph that contains about forty million vertices, we only need about three minutes and 320MB of space to build the indexing graph.

\begin{table}[t]
    \centering\footnotesize
    \caption{The time cost and the memory cost of indexing the alias analysis.}
    \begin{tabular}{c|c|c|c|c|c|c}
    \toprule
         \multirow{2}{*}{\textbf{ID}} & \multicolumn{3}{c|}{\textbf{Time} (seconds)} & \multicolumn{3}{c}{\textbf{Memory} (MB)} \\\cline{2-7}
            & Indexing Graph & PathTree & \textbf{Total}   &  Indexing Graph  &  PathTree  & \textbf{Total}   \\\midrule
        1  &0.09   & 1.80     & 1.89     & 0.21   & 1.08    & 1.29    \\
        2  &0.33   & 8.79     & 9.12     & 0.56   & 2.91    & 3.47    \\
        3  &1.53   & 34.60    & 36.13    & 1.28   & 7.95    & 9.23    \\
        4  &8.00   & 105.67   & 113.67   & 5.13   & 28.76   & 33.89   \\
        5  &3.25   & 41.19    & 44.44    & 3.16   & 16.80   & 19.96   \\
        6  &0.84   & 33.55    & 34.39    & 2.36   & 13.26   & 15.62   \\
        7  &3.26   & 69.85    & 73.11    & 4.07   & 23.97   & 28.04   \\
        8  &1.97   & 83.15    & 85.12    & 8.33   & 38.46   & 46.79   \\
        9  &11.63  & 404.20   & 415.83   & 25.50  & 142.98  & 168.48  \\
        10 &33.85  & 645.69   & 679.54   & 36.00  & 204.34  & 240.34  \\
        11 &37.39  & 1894.22  & 1931.61  & 72.07  & 427.44  & 499.51  \\
        12 &110.80 & 6507.63  & 6618.43  & 141.52 & 925.09  & 1066.61 \\\midrule
        13 &130.52 & 3144.47  & 3274.99  & 173.92 & 890.78  & 1064.70 \\
        14 &196.54 & 10038.21 & 10234.75 & 320.35 & 1763.15 & 2083.50 \\
        15 &124.52 & 3580.17  & 3704.69  & 127.60 & 684.07  & 811.67  \\
        16 &113.56 & 10737.98 & 10851.54 & 297.56 & 1558.47 & 1856.03 \\
        \bottomrule
    \end{tabular}
    \label{tab:flareaa}
\end{table}

\begin{figure}[t]
    \centering
    \includegraphics[width=\textwidth]{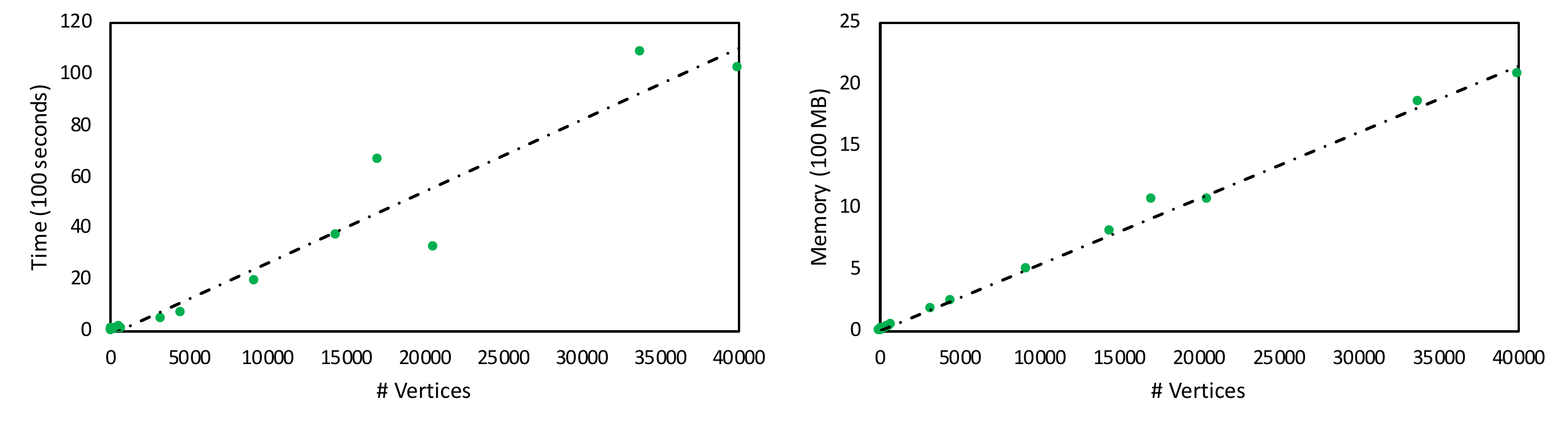}
    \caption{Indexing the alias analysis exhibits linear complexity in practice.}
    \label{fig:flareaa-total}
\end{figure}

Figure~\ref{fig:flareaa-total} shows the total time and space we need to build the index using PathTree (including the cost of building the indexing graph and the cost of computing the PathTree index).
Compared to the Grail indexing scheme in the information-flow analysis,
both the time cost and the memory cost are higher but still tend to be linear as shown in Figure~\ref{fig:flareaa-total}.
For the largest program, it takes less than 3 hours to build the index.
It is noteworthy that the scalability of PathTree has been shown in previous works~\cite{jin2011path} and the PathTree indexing scheme is not our technical contribution.
In practice, if an application cannot afford the overhead of PathTree, we can choose other indexing schemes as discussed in Section~\ref{subsec:gr}.

\begin{figure}[t]
    \centering
    \includegraphics[width=\textwidth]{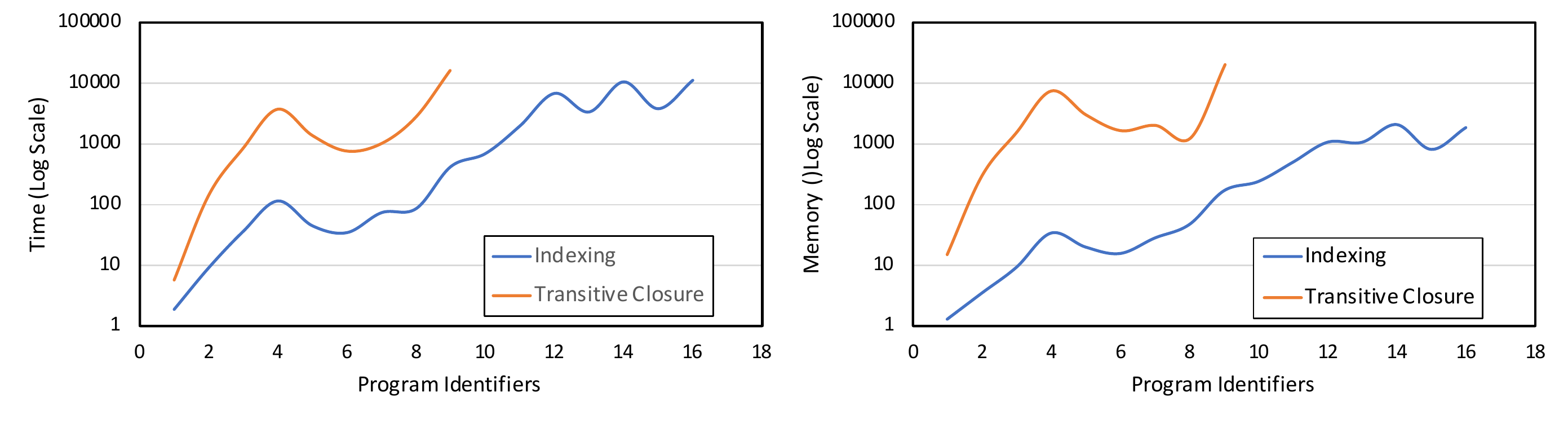}
    \caption{Comparing to computing the transitive closure, the overhead of indexing is reasonable for the context-sensitive alias analysis.}
    \label{fig:flareaa-tc}
\end{figure}

In spite of the high time cost, we show in Figure~\ref{fig:flareaa-tc} that,
in comparison to computing a transitive closure, the overhead of computing the reachability index is much lower and reasonable in practice.
As demonstrated in Figure~\ref{fig:flareaa-tc},
even in a log-scale coordinate system, 
the curves of computing the transitive closure are much higher than those of our approach.
Particularly,
we cannot finish computing the transitive closure for programs with more than four million vertices while we can finish computing the PathTree index for all benchmark programs. Meanwhile, compared to the transitive closures we succeed computing,
the size of our reachability index is much smaller, saving 99.1\% of the space on average.

\subsubsection{Indexing Enables Much Faster Queries.}\label{aa3}
As shown in Table~\ref{tab:flareaa-query}, the reachability index allows 
the indexed alias analysis (\textsc{FlareAA})
to answer 10,000 reachable queries in 200 milliseconds and 10,000 unreachable queries in 30 milliseconds,
81$\times$ to 270,885$\times$ and 134$\times$ to 47,199$\times$ faster than the baseline approach~\cite{li2013precise}.
In total, compared to the baseline approach, the indexing scheme for the alias analysis can speed up the context-sensitive aliasing queries with a speedup from $91\times$ to $248,812\times$, with $5,227\times$ as the median.

In addition to the promising speedup over the conventional approach,
we can observe that the query performance is very stable and works like constant time complexity 
----
the time cost of answering 10,000 reachable and unreachable queries is always around or less than 200 milliseconds and 30 milliseconds, respectively.\footnote{The differences in the query performance between the reachable and the unreachable cases are caused by the internal design of the PathTree indexing scheme, which is not the contribution of this paper and, thus, is omitted.}
This result confirms the impact of our approach, which allows us to answer CS-reachability queries in almost constant time with a moderate space overhead.

\begin{table}[t]
	\centering\footnotesize
	\caption{Time cost (milliseconds) of answering 10,000 reachable queries ($R$), 10,000 unreachable queries ($\neg{R}$), and all queries (Total) in the alias analysis. We also compute the minimum (min), median (med), and maximum (max) speedup compared to the baseline approach.}
	\label{tab:flareaa-query}
	\begin{tabular}{c|r|r|r|r|r|r}
		\toprule
		\multirow{2}{*}{\textbf{ID}} & \multicolumn{3}{c|}{\textbf{\textsc{FlareAA}}} & \multicolumn{3}{c}{\cite{li2013precise}} \\\cline{2-7}
		   & $R$ & $\neg{R}$ & \textbf{Total} & $R$ & $\neg{R}$ & \textbf{Total}\\
		\midrule
1 & 80    & 18       & 99     & 6,532          & 2,454       & 8,985      \\
2 & 102   & 6        & 108    & 83,012         & 21,783      & 104,795    \\
3 & 114   & 12       & 127    & 319,752        & 57,450      & 377,202    \\
4 & 124   & 8        & 133    & 690,446        & 52,262      & 742,708    \\
5 & 108   & 8        & 116    & 243,917        & 38,684      & 282,601    \\
6 & 123   & 8        & 131    & 160,285        & 27,328      & 187,613    \\
7 & 157   & 7        & 164    & 244,014        & 20,068      & 264,082    \\
8 & 149   & 8        & 157    & 289,450        & 34,759      & 324,209    \\
9 & 172   & 11       & 183    & 834,999        & 50,195      & 885,194    \\
10& 155   & 10       & 165    & 3,555,830      & 55,351      & 3,611,181  \\
11& 191   & 8        & 198    & 4,809,880      & 52,284      & 4,862,164  \\
12& 222   & 12       & 234    & 14,419,500     & 122,908     & 14,542,408 \\
13& 196   & 11       & 207    & 7,789,460      & 225,227     & 8,014,687  \\
14& 200   & 10       & 210    & 3,194,340      & 402,685     & 3,597,025  \\
15& 195   & 15       & 210    & 16,580,400     & 402,119     & 16,982,519 \\
16& 216   & 24       & 239    & 58,455,400     & 1,115,120   & 59,570,520 \\
		\midrule\midrule
		\textbf{min} & 81$\times$ & 134$\times$ &  91$\times$   &\multicolumn{3}{c}{}\\
		\textbf{med} & 5,213$\times$ & 5,328$\times$ &  5,227$\times$   &\multicolumn{3}{c}{N/A}\\
		\textbf{max} & 270,885$\times$ & 47,199$\times$ & 248,812$\times$  &\multicolumn{3}{c}{}\\
		\bottomrule
	\end{tabular}
\end{table}

\subsection{Summary} 
According to the evaluation results above, we can come up with two conclusions.
First, the proposed reduction from CS-reachability to conventional graph reachability allows us to benefit from existing indexing schemes to achieve orders of magnitude speedup, at the cost of a reasonable overhead. 
As demonstrated in the evaluation, our approach can scale gracefully for large-scale programs and for a graph with tens of millions of vertices.
	
Second, at the same time of respecting Rule~\ref{p:grail} and Rule~\ref{p:pt}, when choosing an indexing scheme for a program analysis,
we need to consider the overhead induced by the capability of returning paths for each reachable query.
As demonstrated in the evaluation,
such an overhead could make the query slower but still considerably faster than the baseline approach.

\section{Related Work}
\label{sec:relatedwork}

In this section,
we discuss two strands of related work.
Section~\ref{subsec:relwork_cfl} discusses the language-reachability problems in program analysis, particularly for context-sensitive analysis and pointer analysis.
Section~\ref{subsec:relwork_indexing} introduces some advanced reachability indexing schemes with the potential of optimizing program analyses.

\subsection{Language Reachability for Program Analysis}
\label{subsec:relwork_cfl}

Many program analysis problems can be formulated as CFL-reachability problems. Particularly,
Dyck-CFL lays the basis for ``almost all of the applications of CFL reachability in program analysis''~\cite{kodumal2004set}.
Many studies on solving all-pairs CFL or Dyck-CFL reachability have been conducted in various contexts including recursive state machines~\cite{alur2005analysis}, visibly push-down languages~\cite{alur2004visibly}, and streaming XML~\cite{alur2007marrying}.
They usually work with a dynamic programming algorithm, which can be regarded as a generalization of the CYK algorithm for CFL-recognition~\cite{younger1967recognition} and is of cubic time complexity~\cite{kodumal2004set,reps1995precise,yannakakis1990graph}.
\citet{melski2000interconvertibility} and \citet{kodumal2004set} studied the relationship between CFL reachability and set constraint, but their algorithms did not break through the cubic bottleneck.
\citet{chaudhuri2008subcubic} and \citet{zhang2014efficient} showed that the well-known Four Russians' Trick could be employed to achieve sub-cubic algorithms.
Different from the above studies that were conducted in a general setting,
we focus on an extended Dyck-CFL reachability problem for context-sensitive program analysis.
In this setting,
we can easily reduce the CS-reachability problem on a program-valid graph to a conventional reachability problem on the indexing graph,
which allows us to benefit from various indexing schemes from the field of graph databases.
In what follows, we discuss two main applications of language reachability.

\subsubsection{Context-Sensitive Analysis}

\citet{tang2015summary}
employed the tree-adjoining-language (TAL) reachability problem to formulate the context-sensitive data-dependence analysis in the presence of callbacks.
However, their algorithms are of $O(|V|^6)$ time complexity and, thus, are not scalable in practice.
\citet{chatterjee2017optimal} solved the problem by utilizing the ``constant tree-width'' feature of a local data-dependence graph.
However,
their algorithm can only answer a reachability query between vertices in the same calling context.
Our work is different from theirs as we reduce the CS-reachability problem to the conventional reachability problem.

\citet{zhang2017context}
formulated the context-sensitive and field-sensitive data-dependence analysis as a linear-conjunctive-language (LCL) reachability problem.
Compared to the CFL-reachability formulation used in this paper,
LCL-reachability provides a more precise model due to the field sensitivity.
However, 
the exact LCL-reachability problem is known to be undecidable. Thus,
only approximation algorithms can be provided~\cite{reps2000undecidability}.
\citet{li2021complexity} further refined the results of context-sensitive and field-sensitive data-dependence analysis
by formulating it as an interleaved Dyck-CFL reachability problem over a bidirected graph and proved that the interleaved Dyck-CFL reachability problem with more than two parenthesis pairs is NP-hard.
To improve the algorithm efficiency in practice,
\citet{li2020fast} proposed an approach to reducing the graph size before conducting the interleaved Dyck-CFL reachability analysis.

\subsubsection{Pointer Analysis}

The other common use of CFL-reachability in program analysis is to resolve pointer relations on a bidirected graph,
where each edge $(u, v)$ labeled by a left-parenthesis corresponds to an inverse edge $(v, u)$ labeled by a right-parenthesis.
When the underlying language is the Dyck language,
\citet{zhang2013fast} and \citet{chatterjee2017optimal} have demonstrated that a transitive closure can be computed in almost linear time.
However, it is much harder in a general setting. Thus, many techniques have been proposed to optimize the CFL-based pointer analysis.
\citet{zheng2008demand} proposed a demand-driven method to resolve pointer relations without computing the transitive closure.
\citet{xu2009scaling} optimized the CFL-based pointer analysis by computing must-not-alias information for all pairs of variables,
which then can be used to quickly filter out infeasible paths during the more precise pointer analysis.
\citet{zhang2014efficient} optimized the CFL-based pointer analysis by selectively propagating reachability information,
which allows us to bypass a large portion of edges.
\citet{dietrich2015giga} proposed a novel transitive-closure data structure with a
pre-computed set of potentially matching load/store pairs
to accelerate the fix-point calculation.
\citet{wang2017graspan} proposed a graph system that allows CFL-based pointer analysis to work in a single machine by utilizing the disk space.
Our work is different from theirs because we do not focus on bidirected graphs and the underlying context-free language is different.

\subsection{Indexing Schemes and Their Potential Use in Program Analysis}
\label{subsec:relwork_indexing}

Our reduction from the CS-reachability problem to the conventional reachability problem
can be utilized together with other advanced graph database techniques,
thereby enabling more program analysis applications. We briefly discuss some of them below.

\subsubsection{Indexing for Dynamic Graphs}
Beyond the indexing schemes for conventional graph reachability discussed in Section~\ref{sec:preliminaries},
recent studies also consider additional constraints when evaluating reachability queries.
Some of these indexing schemes can be used directly to optimize program analyses.
\citet{roditty2008improved}, \citet{bouros2009evaluating}, and \citet{zhu2014reachability} proposed improved reachability algorithms to handle dynamic graphs,
where edges and vertices may be dynamically created, updated, or deleted.

These indexing schemes, which we refer to as incremental indexing schemes, can be applied to incremental program analysis where program variables or dependence relations are changed during software evolution. When software evolves, we continuously revise the indexing graph proposed in the paper. An incremental indexing scheme can quickly capture the graph changes and rebuild the indexes for answering CS-reachability queries.

\subsubsection{Indexing for Label-Constraint Reachability}
\citet{jin2010computing} proposed the problem of label-constraint reachability (LCR),
in which a vertex $v$
is reachable from a vertex $u$ if and only if there exists a path from the vertex $u$ to the vertex $v$
and the set of the edge labels on the path is a subset of a given label set.
This problem has been extensively studied~\cite{zou2014efficient, valstar2017landmark}.
Recently, \citet{peng2020answering} proposed an indexing scheme to answer billion-scale LCR queries.
\citet{hassan2016graph} and \citet{rice2010graph} proposed approaches to finding the shortest path for LCR problems.

Many program analyses can be modeled as an LCR problem.
For instance,
in a taint analysis,
by modeling the sanitizing operations as edge labels,
we can employ the LCR indexing schemes to check if the tainted data are propagated to a destination with proper sanitizations.
Using the indexing graph proposed in the paper,
we can label its edges with sanitization labels
and use an LCR indexing scheme to enable an efficient context-sensitive taint analysis.

\subsubsection{Benefiting from Other Graph Database Techniques}

Since the indexing graph is a common directed graph,
we can profit from a lot of existing graph database techniques, such as query caching techniques and graph simplification algorithms, to accelerate reachability queries on the indexing graph.
The query caching techniques, such as C-Graph~\cite{zhou2018cgraph}, explore the data locality to speed up graph processing tasks.
All modern graph databases, such as Neo4j and HyperGraphDB,\footnote{Neo4j: \url{https://neo4j.com}; HyperGraphDB: \url{http://www.hypergraphdb.org}.}
implement such caching mechanisms for acceleration.
These approaches are orthogonal to our idea and can be used together with our approach for better performance.

The graph simplification techniques aim to reduce the graph size so as to accelerate reachability queries.
First,
we can follow existing approaches to discover and merge vertices with equivalent reachability relations, e.g., vertices in a strongly connected component or vertices with the same successors and predecessors,
thereby reducing the graph size~\cite{fan2012query,zhou2017dag}.
Second,
we can employ existing advances to perform transitive reduction, which aims to remove unnecessary edges with respect to reachability queries~\cite{zhou2017dag,aho1972transitive,habib1993calculation,simon1988improved,valdes1982recognition,williams2012multiplying}.

\section{Conclusion}
\label{sec:conclusion}

We have presented a reduction from the problem of context-sensitive reachability to the problem of conventional graph reachability. This reduction allows us to efficiently answer context-sensitive reachability queries using the indexing schemes for conventional graph reachability. We apply our approach to speeding up two context-sensitive data flow analyses and compare them with the state-of-the-art approaches. The evaluation results demonstrate that we can achieve orders of magnitude speedup for answering a query, at the cost of only a moderate overhead to build and store the indexes. Since reducing the complexity of CFL-reachability is theoretically very hard in a general setting, providing proper indexing schemes could be a promising solution for reducing the cost of CFL-reachability queries.

\section*{ACKNOWLEDGEMENTS}

We would like to thank the anonymous reviewers and Peisen Yao for their valuable feedback on our earlier paper draft. This work was supported in part by the Ant Research Program from Ant Group and the RGC16206517 and ITS/440/18FP grants from the Hong Kong Research Grant Council.

\bibliographystyle{ACM-Reference-Format}
\bibliography{sigproc}

\appendix
\section{Example of the Exploded super-graph}
\label{app:esg}

\begin{wrapfigure}{R}{0.6\textwidth}
    \begin{center}
        \includegraphics[width=0.59\textwidth]{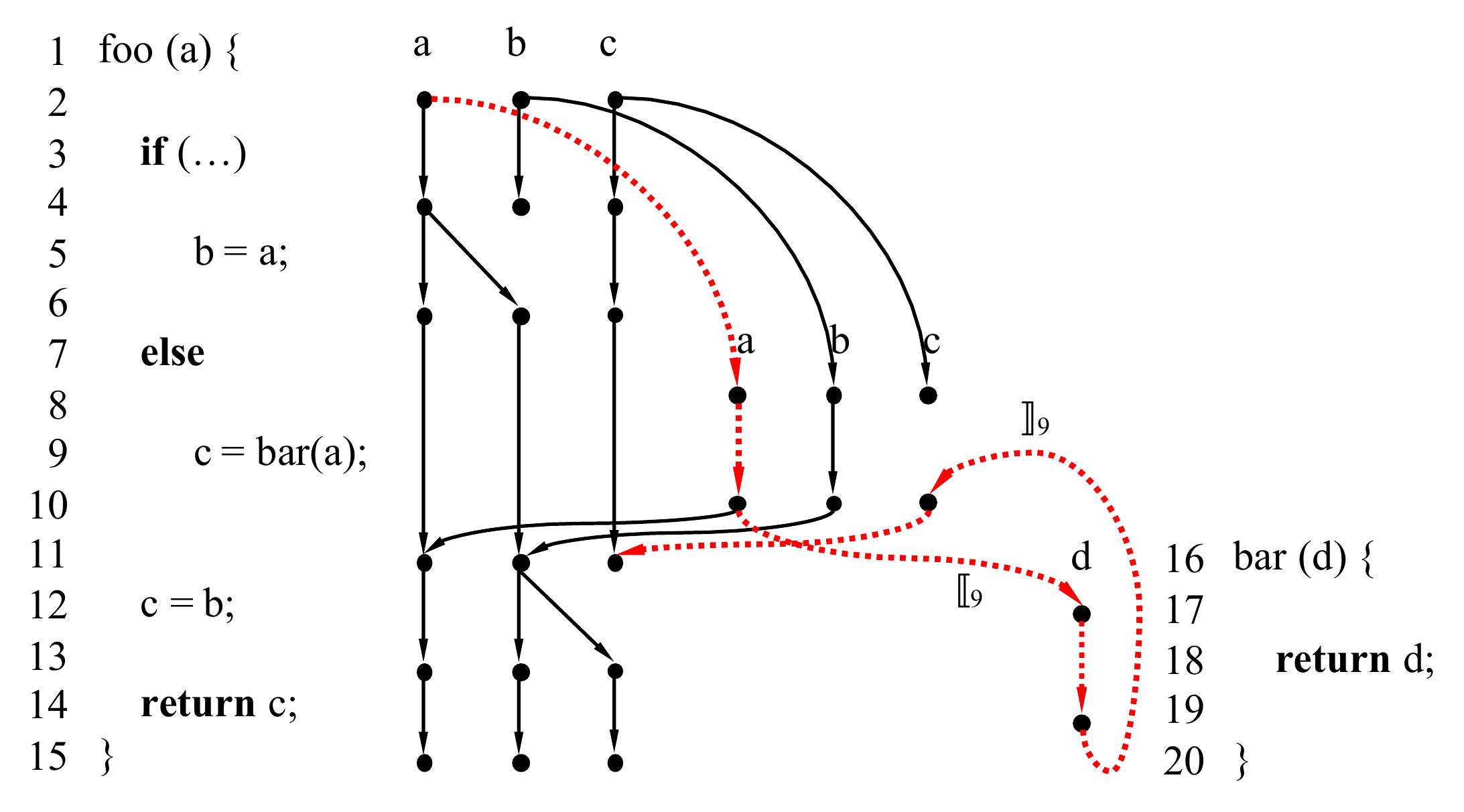}
    \end{center}
    \caption{The program-valid graph, a.k.a., the exploded-super graph, used in the information-flow analysis~\cite{lerch2014flowtwist}. The dotted and red path is a CS-reachable path and represents a valid information flow.}
    \label{fig:app-esg}
\end{wrapfigure}

The IFDS framework solves a wide range of data flow problems with distributive
flow functions over finite domains~\cite{reps1995precise}.
These problems can be reduced to a graph reachability problem over the exploded super-graph.
Figure~\ref{fig:app-esg} demonstrates an example of the exploded super-graph
used in the information-flow analysis~\cite{lerch2014flowtwist}.
A vertex on the exploded-super graph stands for a data flow fact holding at a program point.
For instance,
the vertex $(a, 2)$ means that the data flow fact, i.e., the variable $a$ includes some sensitive information, holds at Line~2.
An edge on the exploded-super graph represents the propagation of data flow facts via a statement.
As an example, the assignment \textit{b = a} will generate a data flow fact that 
the variable $b$ receives some information,
assuming the variable $a$ holds the information before.
Thus, we have an edge from the vertex $(a, 4)$ to the vertex $(b, 6)$.
At the same time,
the data flow fact that the variable $a$ holds the information 
should be kept.
Thus, we have an edge from the vertex $(a, 4)$ to the vertex $(a, 6)$.
At a call site,
we can build the graph edges, namely call edges and return edges, as we reach a series of assignments that
assign the actual parameters to the formal parameters and
assign the return values to the return-value receivers.
Additionally,
we need to add the parentheses, $\llbracket_i$ and $\rrbracket_i$, on the call edges and the return edges to label the call site at Line $i$.
In the example,
the red and dotted path from the vertex $(a, 2)$ to the vertex $(c, 11)$ is a CS-reachable path as the string of the edge labels only includes a pair of matched parentheses, 
thus 
representing a valid information flow.

\section{Example of the Value-Flow Graph}
\label{app:vfg}

\begin{figure}[t]
    \centering
    \includegraphics[width=\textwidth]{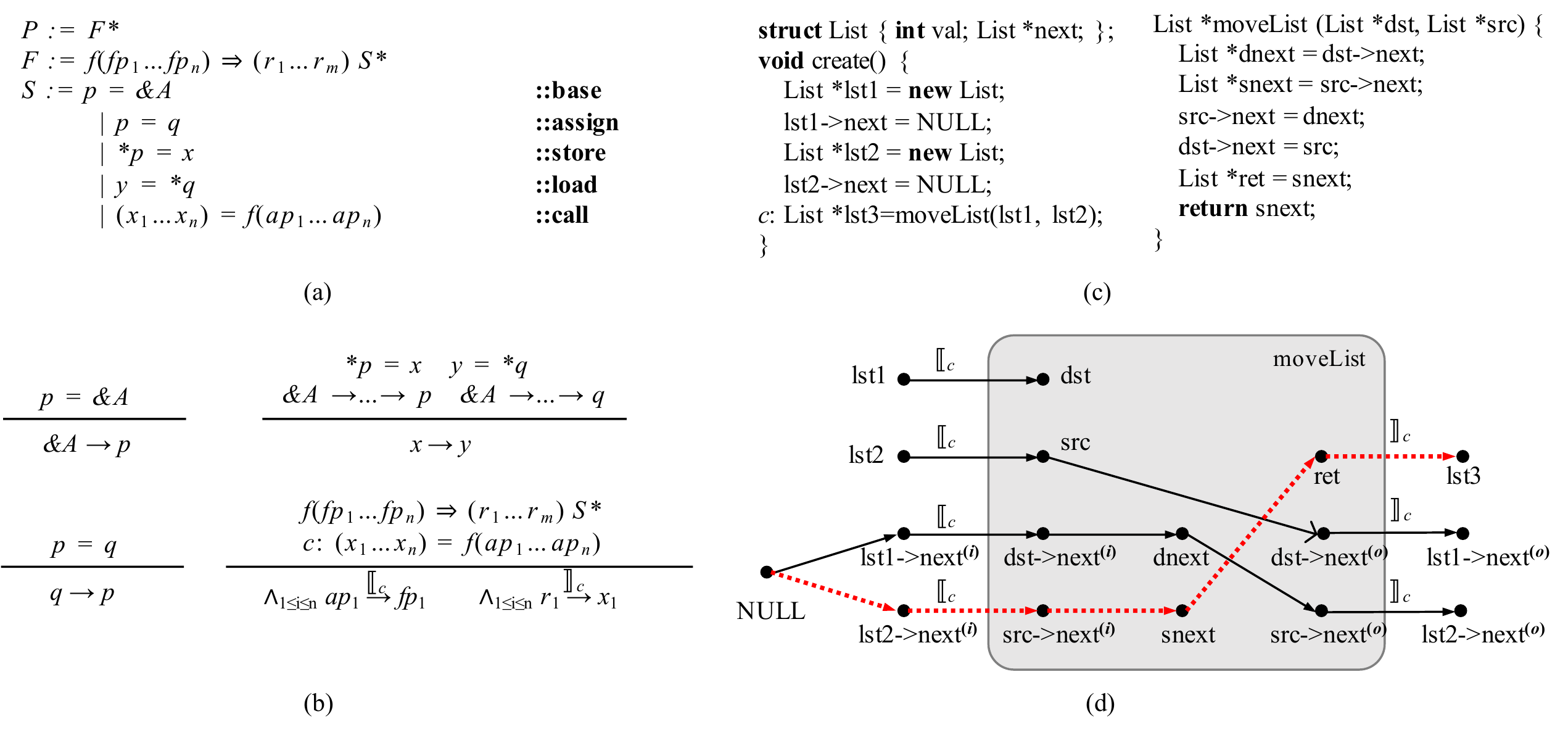}
    \caption{The program-valid graph, a.k.a., the value-flow graph, used in the alias analysis~\cite{li2013precise}. The dotted and red path is a CS-reachable path and all pointers on the path are aliases.}
    \label{fig:app-vfg}
\end{figure}

\citet{li2013precise}'s context-sensitive alias analysis is formulated over the small languages shown in Figure~\ref{fig:app-vfg}(a).
Most statements in the language are standard except that 
every function is assumed to be a pure function (i.e., without any side-effects) and allows multiple formal parameters $\textit{fp}_i$ and multiple returns $\textit{r}_i$.
A function call is represented as $(x_1\dots x_n)=f(\textit{ap}_1\dots\textit{ap}_n)$,
where $x_i$ are return value-receivers and $\textit{ap}_i$ are actual parameters.

The value-flow graph is built based on the rules in Figure~\ref{fig:app-vfg}(b).
For any ``base'' statement $p=\&A$ and ``assign'' statement $p=q$, we add an edge from $\&A$ and $q$ to $p$, respectively.
For a pair of store and load assignments, $*p=x$ and $y=*q$,
we add an edge from $x$ to $y$ if and only if $p$ and $q$ have the same ancestor,
which implies that $p$ and $q$ are pointer aliases.
For a call statement with a label $c$, we build edges by regarding it as a series of assignments that assign the formal parameters to the actual parameters and 
assign the return values to their receivers.
Additionally,
we respectively add the parentheses $\llbracket_c$ and $\rrbracket_c$ on the call edges and the return edges to distinguish different call sites.

Figure~\ref{fig:app-vfg}(d) illustrates the value-flow graph for the C/C++ code snippet in Figure~\ref{fig:app-vfg}(c).
To respect the language in Figure~\ref{fig:app-vfg}(a),
we need to purify the function, \textit{moveList},
by introducing two extra formal parameters to represent \textit{dst->next} and \textit{src->next}, which are referenced in the function.
Also,
we need to introduce two extra return values to represent \textit{dst->next} and \textit{src->next} as they are also modified in the function.
These extra formal parameters and return values are represented as the vertices labeled by $(i)$ and $(o)$ in the value-flow graph.
For instance, the vertex \textit{src->next}$^{(i)}$ stands for the value stored in the memory location \textit{src->next} before the function call.
With the value-flow graph, we can formulate the context-sensitive alias analysis as a CS-reachability problem.
For instance, the dotted and red path is a CS-reachable path and all pointers on the path are aliases.

\end{document}